\documentclass[]{opendatalab}

\usepackage{fontspec}
\usepackage{xeCJK}
\newCJKfontfamily\DianShiCJK[
  Path=fonts/,
  Extension=.otf,
  UprightFont=*
]{NotoSerifCJKsc-Regular}
\usepackage{xcolor}
\usepackage{longtable}
\usepackage{listings}
\usepackage{rotating}
\usepackage{float}
\usepackage[misc]{ifsym}

\newcommand{\dbobject}[1]{\textsf{#1}}
\newcommand{\dbobjectfirst}[1]{{\sffamily\fontseries{bx}\selectfont #1}}
\DeclareFontShape{OT1}{cmss}{b}{n}{<->ssub*cmss/bx/n}{}
\newcommand{\dsReference}{\dbobject{Reference}}
\newcommand{\dsReferences}{\dbobject{References}}
\newcommand{\dsSubstance}{\dbobject{Substance}}
\newcommand{\dsSubstances}{\dbobject{Substances}}
\newcommand{\dsReactionInstance}{\dbobject{Reaction Instance}}
\newcommand{\dsReactionInstances}{\dbobject{Reaction Instances}}
\newcommand{\dsReactionGroup}{\dbobject{Reaction Group}}
\newcommand{\dsReactionGroups}{\dbobject{Reaction Groups}}
\newcommand{\dsReactionTemplate}{\dbobject{Reaction Template}}
\newcommand{\dsReactionTemplates}{\dbobject{Reaction Templates}}

\title{DianShi-RxnDB: A Large-Scale, Fine-Grained Organic Reaction Data Platform Built via a Fully Automated Pipeline for Researchers and AI Agents}

\author[1\dag]{Yubin Wang}
\author[1\dag]{Xingjian Wei}
\author[1\dag\ddagger]{Jiang Wu}

\author[1]{Yinfan Wang}
\author[2]{Boyu Zhu}

\author[1]{Lin Zhang}
\author[1]{Jianing Yu}

\author[2]{Huazheng Zeng}
\author[2]{Ruiyi Ding}
\author[1]{Junyuan Gao}
\author[1]{Jiaxing Sun}
\author[3]{Lingli Ge}

\author[1]{Haote Yang}
\author[4]{Jingchao Wang}

\author[1]{Aijia Guo}
\author[1]{Qian Jiang}

\author[1]{Yurui Zhao}
\author[1]{Wenjian Zhang}
\author[5]{Chen Zhu}
\author[1]{Lijun Wu}

\author[1]{Xiaolei Yang}
\author[1]{Haodong Chen}
\author[1]{Junjie Yuan}

\author[1]{Zichao Ye}
\author[1]{Shaowei Hou}

\author[1]{Jing Ye}
\author[1]{Jia Yu}
\author[1]{Shan Wang}
\author[1]{Lijun Wu}
\author[1]{Jiantao Qiu}
\author[1]{Chao Xu}
\author[1]{Yuqiang Li}
\author[1]{Guangyu Wang}

\author[1]{Bowen Zhou}
\author[1]{Dahua Lin}
\author[1\ \textrm{\Letter}]{Conghui He}

\affiliation[1]{Shanghai Artificial Intelligence Laboratory}
\affiliation[2]{Fudan University}
\affiliation[3]{Shanghai Jiao Tong University}
\affiliation[4]{East China Normal University}
\affiliation[5]{East China University of Science and Technology}

\abstract{%
High-quality structured organic reaction data underpin reaction-precedent retrieval, investigation of reported experimental conditions, and applications in artificial intelligence for chemistry (AI4Chem).
Much of the relevant synthetic knowledge, however, is dispersed across the text, images, and reaction schemes of patent documents, making it difficult to search, compare, and process computationally.
We present DianShi-RxnDB, a large-scale, fine-grained organic reaction data platform for organic chemistry researchers and AI agents.
The platform is built through a fully automated information-extraction and normalization pipeline that covers patent text, images, and reaction schemes.
Its patent corpus is drawn from the United States Patent and Trademark Office (USPTO) and the European Patent Office (EPO) and primarily covers organic synthesis patents published between 1976 and 2025.
The database contains approximately 24 million \dsReactionInstances{}, of which approximately 14.8 million (61.7\%) pass the implemented automated qualification checks.
The database organizes reaction knowledge as specific single-step reaction records extracted from patent documents.
Each \dsReactionInstance{} records reaction participants and their roles, quantities, temperatures, times, yields, and experimental procedures, and is linked to its source patent and relevant source location.
These structured relationships support the retrieval, comparison, and source verification of related experiments.
We randomly sampled 1,300 records from the qualified-instance population for manual quality evaluation; across yield, reactant, reagent, catalyst, and solvent fields, the micro-averaged field-level accuracy was 92.95\%.
An external matched comparison with Pistachio further showed advantages for DianShi-RxnDB in the evaluated dimensions, including reaction-record counts after deduplication, representation granularity, and field-level exact agreement against source-grounded references.
On the same data foundation, the Web research workbench supports researchers with search, filtering, comparison of single-step reaction records, linked exploration, and source-patent verification, while the Model Context Protocol (MCP) service provides AI agents with composable structured retrieval tools for multi-step queries and result organization.
Users can access DianShi-RxnDB through the Web research workbench at
\textbf{\href{https://dianshi.opendatalab.org.cn/}
{https://dianshi.opendatalab.org.cn/}}
and connect it to AI agents through the MCP service at
\textbf{\href{https://dianshi.opendatalab.org.cn/mcp}
{https://dianshi.opendatalab.org.cn/mcp}}.

}

\date{\today}
\metadata[Equal Contribution ($\dagger$)]{Yubin Wang, Xingjian Wei, Jiang Wu}
\metadata[Project Lead ($\ddagger$)]{Jiang Wu, \email{wujiang@pjlab.org.cn}}
\correspondence{Conghui He, \email{heconghui@pjlab.org.cn}}

\begin{document}

\maketitle

\begin{tcolorbox}[colback=odlbg,colframe=odlblue,boxrule=0.5pt,arc=1pt,left=8pt,right=8pt,top=6pt,bottom=6pt]
\small\itshape DianShi, the romanized Chinese name {\DianShiCJK 点石}, evokes the idiom “turning stone into gold” and represents our mission to transform organic-chemistry knowledge scattered across patents, papers, and other literature into structured, searchable, AI-ready data and intelligence.
\end{tcolorbox}

\begin{figure*}[t]
\centering
\includegraphics[width=\textwidth,keepaspectratio]{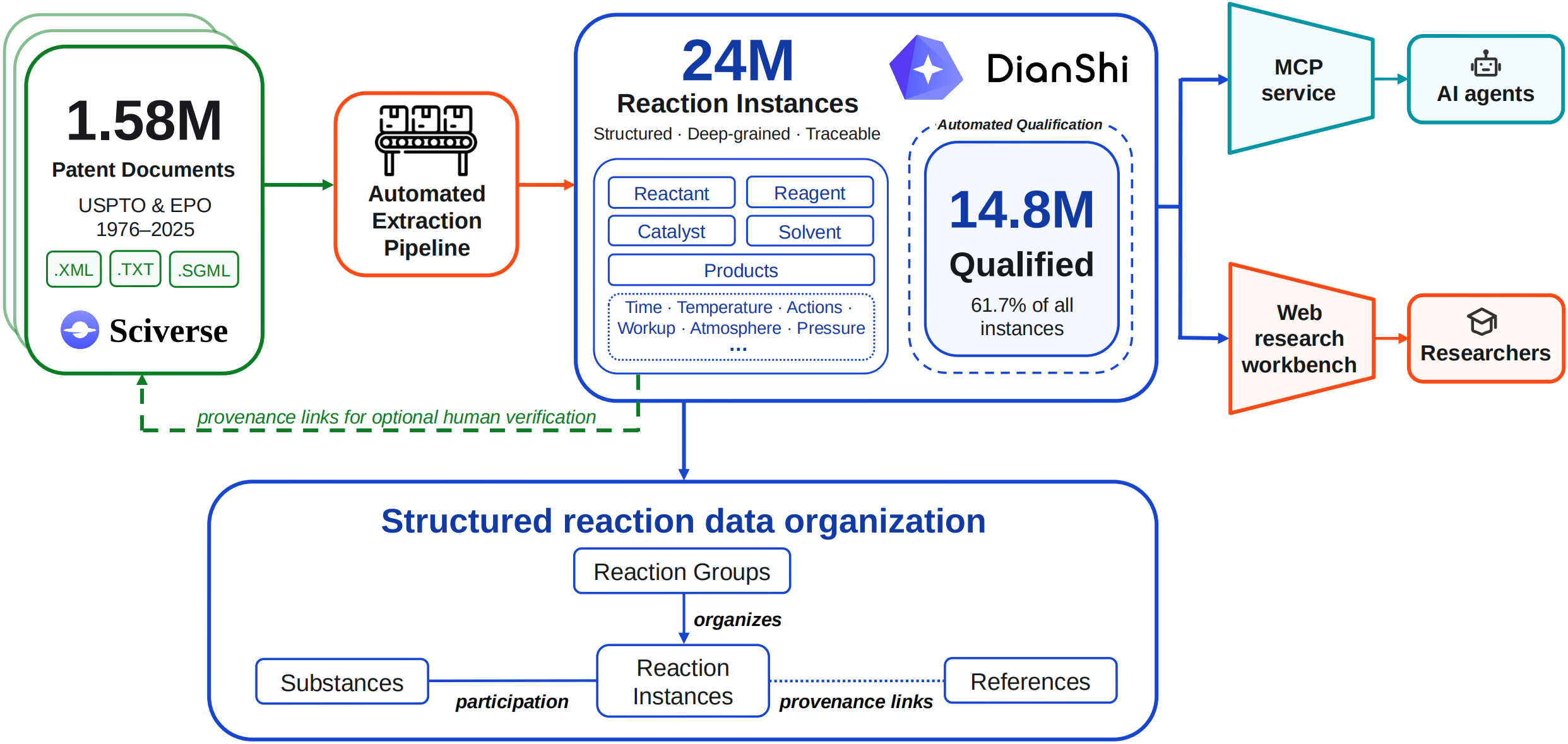}
\caption{System architecture and data-flow overview of DianShi-RxnDB.
Through a fully automated information-extraction and normalization pipeline, DianShi-RxnDB transforms approximately 1.58 million patent documents from the USPTO and EPO, primarily covering 1976--2025, into approximately 24 million fine-grained, provenance-linked \dsReactionInstances{}, of which approximately 14.8 million (61.7\%) pass automated qualification assessment.
The instance-level data include reaction participants, experimental conditions, and procedures, with structured relationships among \dsSubstances{}, \dsReactionInstances{}, \dsReactionGroups{}, and \dsReferences{}.
These relationships support the organization and comparison of related single-step reaction records and provide a path back to source patents for verification.
On this shared data foundation, the Web research workbench provides interactive search, comparison, and source verification for researchers, while the MCP service provides AI agents with structured, composable, multi-step retrieval capabilities.}
\label{fig:system-overview}
\end{figure*}

\section{Introduction}
\label{sec:introduction}

\subsection{Background and existing organic reaction data resources}
\label{sec:background-resources}

Organic synthesis is central to the discovery and development of pharmaceuticals, materials, agrochemicals, and other fine chemicals.
For reaction-precedent retrieval and AI4Chem research, a useful reaction record should describe not only the structural changes between reactants and products but also experimental details such as participant roles, quantities, conditions, yield, and procedure~\citep{kearnes2021ord}.
Such single-step reaction records support the retrieval and comparison of reported experimental procedures and conditions, while providing data for reaction-outcome prediction~\citep{schwaller2019molecular,coley2017outcomes}, reaction-condition prediction~\citep{gao2018conditions}, and data-driven retrosynthetic planning~\citep{segler2018planning,thakkar2020datasets}.

Much organic reaction knowledge remains distributed across journal articles and patents and cannot be directly converted into uniform, machine-readable reaction records~\citep{lowe2012thesis,guo2022reactionextraction}.
Reaction structures, experimental conditions, yields, and procedures may be distributed across the main text, reaction schemes, tables, captions, experimental sections, or supporting information, requiring text, tables, and images to be interpreted together for complete extraction~\citep{vaucher2020actions,fan2024openchemie}.
Patent documents are particularly complex: information about a specific experiment may span different paragraphs, reaction steps, and examples, while some compounds and operations can be understood accurately only by combining general procedures, context, and in-document cross-references~\citep{lowe2012thesis}.
Forming fine-grained reaction data from primary documents therefore requires not only recognition of chemical entities and reaction relationships but also experimental-boundary determination, contextual linking, source localization, and representational normalization~\citep{lowe2012thesis,guo2022reactionextraction,hawizy2011chemicaltagger}.
Scaling these processes to large document collections remains a central challenge in building structured reaction data.

Existing organic reaction resources adopt diverse forms of organization and service.
Public reaction datasets are commonly released as downloadable files or through data repositories, providing useful foundations for cheminformatics and AI4Chem research.
Professional databases, in contrast, curate chemical information from journals, patents, and other sources and provide search, retrieval, and linked-exploration services.
\Cref{tab:reaction-resources} compares representative resources in terms of reported scale, source-literature coverage, reaction and experimental information, provenance localization, researcher-facing access, machine-access options, and access conditions.

\begin{table}[H]
\centering
\scriptsize
\setlength{\tabcolsep}{2pt}
\renewcommand{\arraystretch}{1.15}
\caption{
Representative organic reaction data resources and their system characteristics.
The table compares the reported scale, original-literature coverage, reaction and experimental fields, provenance location, and access mode of each resource.
Here, ``reaction and experimental fields'' refers to information captured at the reaction-record or reaction-instance level, including participant roles, quantities or scale, conditions, multistage support, structured procedures, yield, and other outcomes.
The symbols indicate fields or capabilities documented in the cited materials and do not imply that every record contains every field.
For the field notation, $\mathcal{R}$ denotes reaction participants and roles, $\mathcal{Q}$ quantities or scale, $\mathcal{C}$ reaction conditions, $\mathcal{S}$ support for multistage reactions, $\mathcal{P}$ a structured procedure mapped from the literature to predefined operations and parameters, $\mathcal{Y}$ yield, and $\mathcal{O}$ outcomes other than yield.
For provenance location, ``document'' indicates an association with the original literature without confirmed paragraph-level localization, whereas ``paragraph'' indicates a paragraph-level location.
}
\label{tab:reaction-resources}
\newcommand{\tablehead}[2]{\shortstack[l]{\strut #1\\\strut #2}}
\resizebox{\linewidth}{!}{%
\begin{tabular}{>{\raggedright\arraybackslash}p{2.8cm}>{\raggedright\arraybackslash}p{1.4cm}>{\raggedright\arraybackslash}p{1.6cm}>{\raggedright\arraybackslash}p{3.1cm}>{\raggedright\arraybackslash}p{1.8cm}>{\raggedright\arraybackslash}p{4.0cm}}
\toprule
\tablehead{Resource}{name} &
\tablehead{Reported}{scale} &
\tablehead{Literature}{coverage} &
\tablehead{Reaction and}{experimental fields} &
\tablehead{Provenance}{location} &
\tablehead{Access}{mode} \\
\midrule
\mbox{USPTO-Lowe~\citep{lowe2017uspto,schneider2016patents}} &
\mbox{\(\sim\)3.75~M} &
\mbox{1976--2016} &
\mbox{$\mathcal{R} \!\cdot\! \mathcal{Q} \!\cdot\! \mathcal{C} \!\cdot\! \mathcal{P} \!\cdot\! \mathcal{Y}$} &
\mbox{Document} &
\mbox{Public download} \\
\mbox{CJHIF~\citep{jiang2021smiles,jiang2021data}} &
\mbox{\(\sim\)3.22~M} &
\mbox{Not specified} &
\mbox{$\mathcal{R} \!\cdot\! \mathcal{Y}$} &
\mbox{Not localized} &
\mbox{Public repository} \\
\mbox{USPTO-LLM~\citep{yuan2025usptollm,yuan2024zenodo,gong2024usptollm}} &
\mbox{\(\sim\)247,000} &
\mbox{1976--2016} &
\mbox{$\mathcal{R} \!\cdot\! \mathcal{C} \!\cdot\! \mathcal{S}$} &
\mbox{Document} &
\mbox{Public download} \\
\mbox{Pistachio~\citep{pistachio2026,mayfield2017pistachio,mayfield2021pistachio}} &
\mbox{\(\sim\)21~M} &
\mbox{1971--present} &
\mbox{$\mathcal{R} \!\cdot\! \mathcal{Q} \!\cdot\! \mathcal{C} \!\cdot\! \mathcal{P} \!\cdot\! \mathcal{Y}$} &
\mbox{Paragraph} &
\mbox{Paid Web access} \\
\mbox{Reaxys~\citep{reaxysAcademic,reaxysProduct}} &
\mbox{\(\sim\)73~M} &
\mbox{1771--present} &
\mbox{$\mathcal{R} \!\cdot\! \mathcal{C} \!\cdot\! \mathcal{Y}$} &
\mbox{Document} &
\mbox{Paid Web; MCP/API license} \\
\mbox{CAS Reactions~\citep{casReactions,casSciFinder}} &
\mbox{\(\sim\)150~M} &
\mbox{1840--present} &
\mbox{$\mathcal{R} \!\cdot\! \mathcal{Q} \!\cdot\! \mathcal{C} \!\cdot\! \mathcal{S} \!\cdot\! \mathcal{Y}$} &
\mbox{Document} &
\mbox{Paid Web access} \\
\mbox{\textbf{DianShi-RxnDB}} &
\mbox{\textbf{\(\sim\)24~M}} &
\mbox{\textbf{1976--2025}} &
\mbox{$\boldsymbol{\mathcal{R}} \!\cdot\! \boldsymbol{\mathcal{Q}} \!\cdot\! \boldsymbol{\mathcal{C}} \!\cdot\! \boldsymbol{\mathcal{S}} \!\cdot\! \boldsymbol{\mathcal{P}} \!\cdot\! \boldsymbol{\mathcal{Y}} \!\cdot\! \boldsymbol{\mathcal{O}}$} &
\mbox{\textbf{Paragraph}} &
\mbox{\textbf{Free non-commercial Web/MCP access}} \\
\bottomrule
\end{tabular}%
}
\end{table}

However, existing open reaction datasets have substantial limitations in data scale, patent-literature coverage, instance-level experimental information, provenance localization, and consistency of data organization.
They cannot provide a complete, reliable, and structurally consistent data foundation for fine-grained reaction retrieval, experimental-condition comparison, source verification, and AI4Chem applications~\citep{hasic2026consolidation}.
Professional databases provide richer curated information and retrieval capabilities, but typically require paid subscriptions or commercial licenses, while machine access, batch use, and system integration may also be subject to corresponding restrictions.

AI is creating new technical and usage requirements for reaction-data platforms.
On the data-construction side, advances in language models and chemical-information processing make it increasingly practical to extract and normalize reaction information from the text, images, and reaction schemes of large document collections.
On the usage side, AI agents require structured and composable retrieval tools that can support multi-step queries while preserving links to the underlying records and source documents.
A useful platform for this setting therefore needs to combine large-scale and fine-grained reaction data, instance-level experimental detail, localized provenance for source verification, and complementary interfaces for researchers and AI agents.
DianShi-RxnDB is designed around this combination of capabilities.

\subsection{Overview of the DianShi-RxnDB}
\label{sec:dianshi-overview}

DianShi-RxnDB is a large-scale, fine-grained organic reaction data platform for organic chemistry researchers and AI agents.
Its data foundation is built through a fully automated information-extraction and normalization pipeline covering patent text, images, and reaction schemes.
The underlying patent documents are drawn from the USPTO and EPO and primarily cover organic synthesis patents published from 1976 through 2025.
The database contains approximately 24 million \dsReactionInstances{}, of which approximately 14.8 million (61.7\%) pass the automated qualification assessment.
The report further presents an external matched comparison with the Pistachio Reaction Dataset, examining reaction-record counts after deduplication, representation granularity, and field-level exact agreement against source-grounded references.

DianShi-RxnDB organizes reaction knowledge as specific single-step reaction records extracted from patent documents.
Each \dsReactionInstance{} records the structural representations of reactants and products, participant roles, per-substance amounts and equivalents, temperature, time, yield, and experimental procedure, and is linked to its source patent and relevant source location.
The database further establishes structured relationships among \dsSubstance{}, \dsReactionInstance{}, \dsReactionGroup{}, and \dsReference{} objects.
A \dsReactionGroup{} organizes multiple single-step reaction records with the same reactant--product combination, while retaining the conditions, yield, procedure, and source information of each instance, thereby supporting the retrieval, comparison, and source verification of related records.

On the same structured reaction data foundation, DianShi-RxnDB provides two complementary interfaces for researchers and AI agents.
The Web research workbench supports search, filtering, instance comparison, linked exploration, and source verification for researchers, while the Model Context Protocol (MCP)~\citep{mcp2026spec} service organizes query and retrieval capabilities for substances, reactions, and source documents as composable structured tools, supporting AI agents in conducting multi-step retrieval and organizing relevant results.
Researchers can use the chemical representations, object identifiers, and source information returned by MCP to locate corresponding records in the Web research workbench and further verify the source patents.

\subsection{Contributions and organization of this report}
\label{sec:report-organization}

This report introduces DianShi-RxnDB from three perspectives: data construction, knowledge organization, and system services.

\begin{enumerate}
\item \textbf{A large-scale, fine-grained reaction data foundation built through a fully automated pipeline.}
The platform uses a fully automated information-extraction and normalization pipeline covering patent text, images, and reaction schemes to construct approximately 24 million \dsReactionInstances{} from organic synthesis patents from the USPTO and EPO, of which approximately 14.8 million pass the automated qualification assessment.
We describe the data sources, construction scope, and capability-level process, and report a manual quality evaluation of five core fields based on a random sample of 1,300 qualified instances, with a micro-averaged accuracy of 92.95\% (\Cref{sec:data-foundation-quality}).
We further present an external matched comparison with the Pistachio Reaction Dataset, in which DianShi-RxnDB retained more reaction records after deduplication, showed finer-grained information organization in the inspected representative record, and achieved higher field-level exact agreement against source-grounded references across all six evaluated fields (\Cref{sec:pistachio-comparison}).
\item \textbf{Instance-level organization of reaction knowledge and provenance linkage.}
The database records participant roles, quantities, reaction conditions, yields, and experimental procedures for specific single-step reaction records, and establishes structured relationships among \dsSubstance{}, \dsReactionInstance{}, \dsReactionGroup{}, and \dsReference{} objects.
These relationships support the comparison of related experiments and verification against their source patents (\Cref{sec:data-foundation-quality,sec:web-workbench}).
\item \textbf{Complementary researcher and AI-agent interfaces over the same data foundation.}
The Web research workbench supports researchers in interactive search, filtering, instance comparison, linked exploration, and source verification, while the MCP service provides AI agents with composable structured tools for multi-step retrieval and result organization.
We demonstrate the two interfaces and their complementary use in the same reaction-precedent retrieval task (\Cref{sec:web-workbench,sec:mcp-service,sec:use-cases}).
\end{enumerate}

\Cref{sec:limitations-responsible-use} discusses data and use boundaries, risks associated with agent use, and availability, while \Cref{sec:conclusion} concludes the report and looks ahead to future work.

\section{Data foundation, construction, and quality evaluation}
\label{sec:data-foundation-quality}

\subsection{Data sources and construction scope}
\label{sec:data-sources}

The source patent records used in constructing the DianShi-RxnDB are drawn from data resources aggregated by the Sciverse scientific data infrastructure~\citep{sciverse2026scibase}; the corresponding patent documents come from the United States Patent and Trademark Office (USPTO) and the European Patent Office (EPO), primarily covering organic synthesis patents published from 1976 through 2025.
Database construction began with a large pool of source patent records and applied domain filtering, record consolidation and deduplication, and full-text availability checks to form the corpus used for reaction information extraction.

The initial collection contained 20,256,438 patent records before deduplication, comprising 12,377,492 USPTO records and 7,878,946 EPO records.
\Cref{tab:patent-processing-stages} summarizes the successive processing stages.

\begin{table}[t]
\centering
\small
\caption{Processing stages from source patent records to the reaction-extraction corpus.}
\label{tab:patent-processing-stages}
\begin{tabular}{p{5.2cm}rrr}
\toprule
Processing stage &
Input records &
Removed records &
Retained records \\
\midrule
Source USPTO and EPO patent records &
--- &
--- &
20,256,438 \\
IPC-based filtering for organic chemistry relevance &
20,256,438 &
17,743,513 &
2,512,925 \\
Within-source consolidation and deduplication &
2,512,925 &
544,695 &
1,968,230 \\
Cross-source deduplication between USPTO and EPO &
1,968,230 &
210,475 &
1,757,755 \\
Removal of records without usable full text &
1,757,755 &
176,816 &
1,580,939 \\
\bottomrule
\end{tabular}
\end{table}

Within-source consolidation merges duplicate application records and different publication versions associated with the same source.
Cross-source deduplication identifies records represented in both the USPTO and EPO collections.
Records were retained conservatively when the available metadata did not support a reliable duplicate determination.

After these stages, 1,580,939 patent documents entered the reaction information extraction pipeline.
Of these documents, 608,309 ultimately link to at least one \dsReactionInstance{} retained in the database.

Further details on the patent-record processing stages and the different patent-counting conventions are provided in \Cref{app:patent-processing-counts}.

\subsection{From patent documents to structured \dsReactionInstances{}}
\label{sec:reaction-instance-construction}

The DianShi-RxnDB produces structured \dsReactionInstances{} from the patent documents in the reaction-extraction corpus through a fully automated information-extraction and normalization pipeline.
The pipeline automatically identifies content related to organic synthesis experiments in patent text and parsable images and reaction schemes, without requiring manual extraction and organization of individual records, and organizes the participants, experimental conditions, yield, experimental procedure, and source location of each specific experiment into a structured \dsReactionInstance{}.
The manual quality evaluation described in \Cref{sec:manual-quality-evaluation} assesses the quality of the pipeline outputs and does not participate in the record-by-record production of \dsReactionInstances{}.

At the capability level, the pipeline covers the parsing of patent text, images, and reaction schemes; the identification and organization of organic-synthesis experimental content; the structuring of reaction participants, experimental conditions, yields, and experimental procedures; and the normalization of chemical representations and experimental fields.
The resulting \dsReactionInstances{} are linked to source patents and relevant source-text locations and are further subjected to automated qualification assessment.
Beyond reactant--product representations, each \dsReactionInstance{} retains instance-level experimental information and its provenance relationship, supporting subsequent retrieval, comparison, and source verification.
For reaction participants with usable structural information, the system uses canonical SMILES to represent their molecular structures.
This report presents only a capability-level overview of the extraction pipeline; its specific technical details will be described in a subsequent technical report.

In the data production described in this report, DeepSeek-V3-0324 is used primarily to process patent experimental text~\citep{deepseek2025v30324}, while MinerU.Chem is used primarily to parse chemical information in images and reaction schemes~\citep{yang2026mineruchem}.
DeepSeek-V3-0324 is deployed on Huawei Ascend 910C AI processors, with DeepLink providing software--hardware adaptation and inference-runtime support~\citep{deeplink2026}.

Each \dsReactionInstance{} records participants and their roles, including Reactant, Product, Reagent, Solvent, and Catalyst.
It also retains available participant quantities and equivalents, experimental conditions such as temperature and time, yield, and experimental procedures.
Each \dsReactionInstance{} is linked to its source patent and the corresponding source-text location or region, allowing users to return from a structured record to the source document and verify relevant experimental content.
Each \dsReactionInstance{} can organize the information categories listed in \Cref{tab:reaction-instance-information}, although the availability of individual fields depends on the content of the source patent and the extraction result.

\begin{table}[t]
\centering
\small
\caption{Principal information organized in a \dsReactionInstance{}.}
\label{tab:reaction-instance-information}
\begin{tabular}{p{3.6cm}p{9.0cm}}
\toprule
Information category &
Principal content \\
\midrule
Reaction representation &
Reactant--product structural representation; Reaction SMILES \\
Participants and roles &
Reactant, Product, Reagent, Catalyst, and Solvent \\
Instance-level amounts &
Available information such as quantities, equivalents, and product amounts \\
Reaction conditions &
Available conditions such as temperature and time \\
Outcome information &
Experimental results such as yield \\
Experimental process &
Available experimental procedures and workup descriptions \\
Provenance linkage &
Source patent, available in-document location, and relevant source content \\
\bottomrule
\end{tabular}
\end{table}

Examples of the principal \dsReactionInstance{} fields are provided in \Cref{app:reaction-instance-fields}; operational definitions used for structured reaction-process details are listed in \Cref{app:reaction-operation-definitions}.

The system applies an automated qualification assessment to generated \dsReactionInstances{}.
This assessment uses RXNMapper~\citep{schwaller2021rxnmapper} to generate atom mappings and checks whether the mapped reaction passes atom-conservation checks and other implemented rules.
\dsReactionInstances{} that pass this assessment are termed \emph{qualified \dsReactionInstances{}}, or \emph{qualified instances} for short.

The database also retains \dsReactionInstances{} that do not pass the automated qualification assessment, together with their extracted experimental information and provenance relationships.
The counts and proportions of the two categories are reported in \Cref{sec:database-scale}.

\begin{figure*}[t]
\centering
\includegraphics[width=\textwidth]{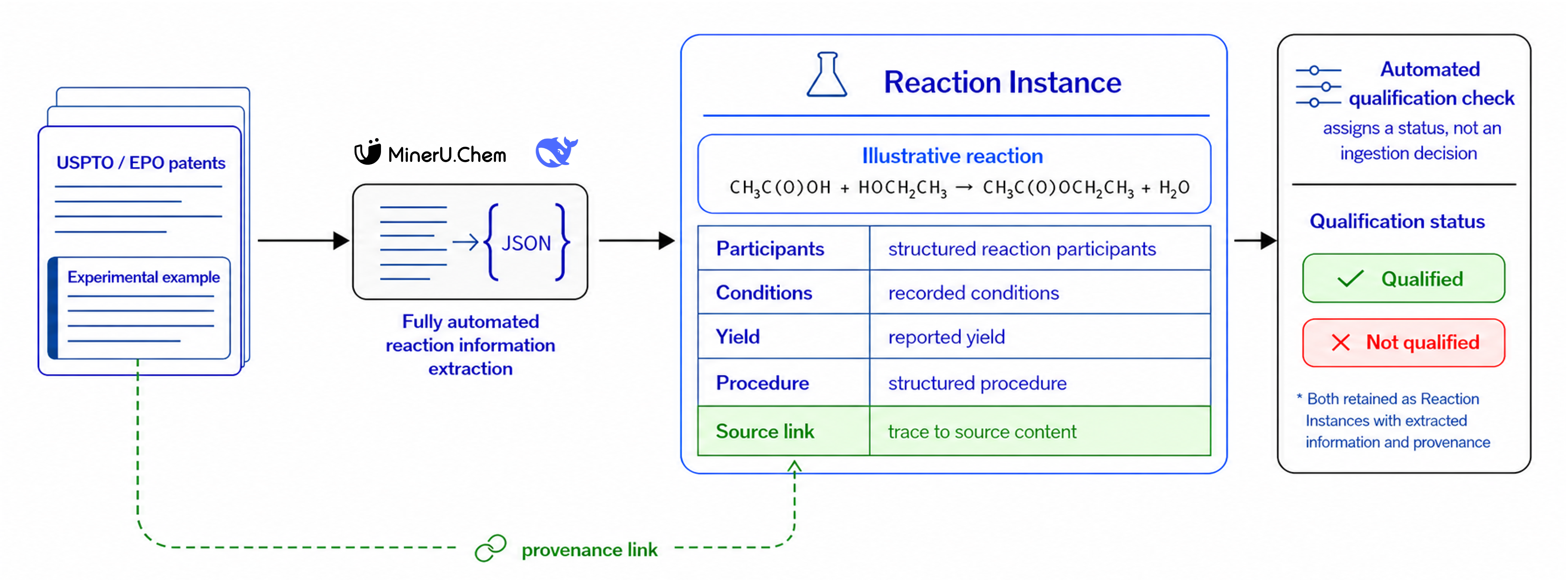}
\caption{DianShi-RxnDB transforms patent documents into provenance-linked \dsReactionInstances{} through a fully automated information-extraction and normalization pipeline.
The pipeline parses patent text, images, and reaction schemes and, without manual extraction and curation of individual records, identifies organic synthesis experiments and organizes reaction participants, experimental conditions, yields, and experimental procedures into structured \dsReactionInstances{}; chemical information in images and reaction schemes is primarily parsed by MinerU.Chem.
Each instance retains a link to the relevant source-patent location to support source-text verification.
Automated qualification then assigns each generated record a ``qualified'' or ``not passed'' status.
Both categories are retained in the database with their extracted information and provenance links.}
\label{fig:construction-provenance-flow}
\end{figure*}

The capability-level data-formation process from patent documents to \dsReactionInstances{} is summarized in \Cref{fig:construction-provenance-flow}.

\subsection{Database objects and relationships}
\label{sec:database-objects}

\begin{figure*}[t]
\centering
\includegraphics[width=\textwidth]{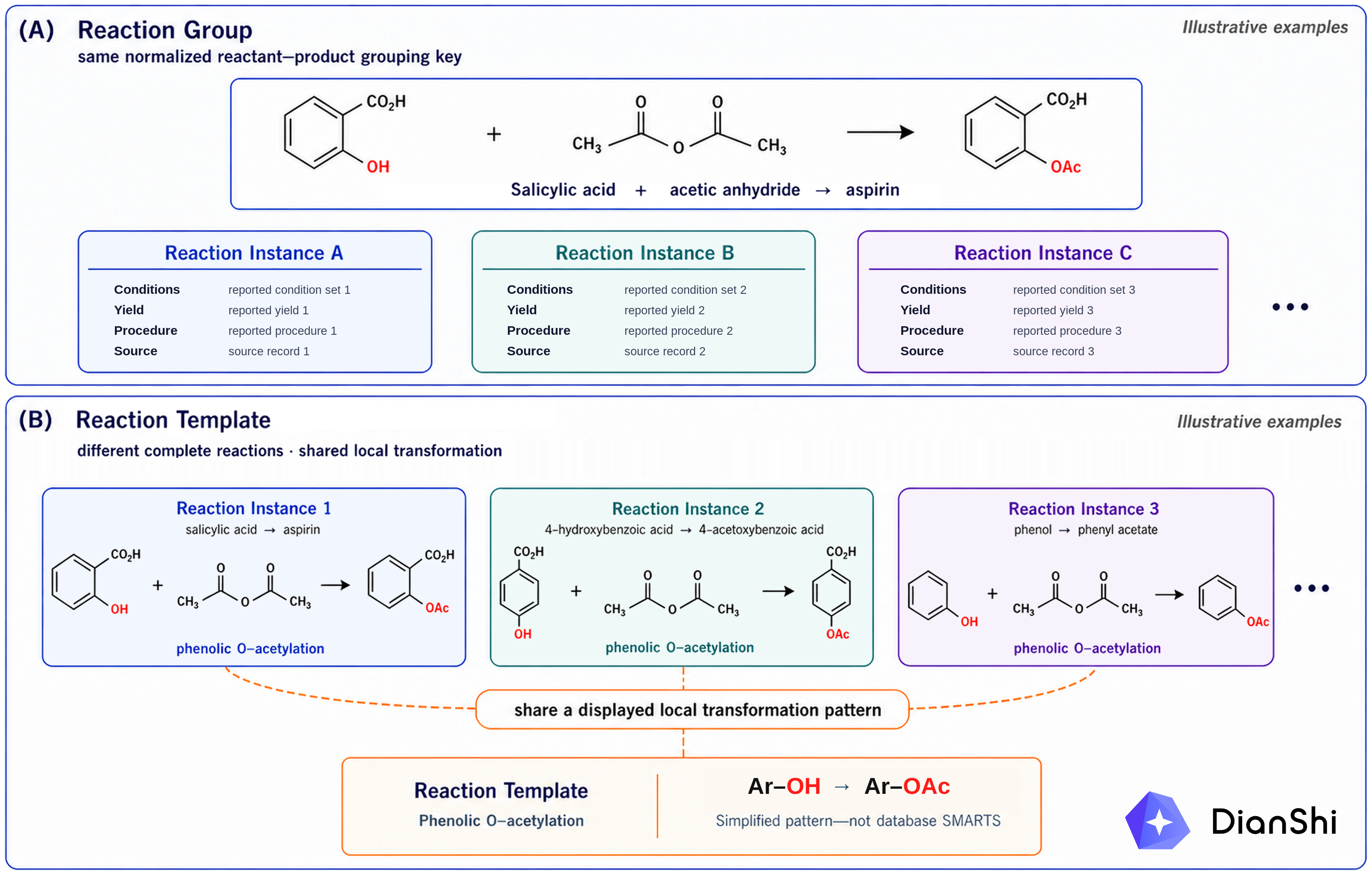}
\caption{Conceptual relationships among \dsReactionInstances{}, \dsReactionGroups{}, and \dsReactionTemplates{}.
(A) A \dsReactionInstance{} represents a specific reported single-step reaction and independently retains its reaction conditions, yield, experimental procedure, and source information.
A \dsReactionGroup{} organizes multiple \dsReactionInstances{} using the same normalized reactant--product grouping key, without merging their experimental information.
(B) A \dsReactionTemplate{} represents a local structural transformation pattern shared by different complete reaction instances, so instances with different specific reactant--product combinations may be associated with the same template.
The local transformation shown in the figure is a simplified pattern for illustration and is not the SMARTS expression actually stored by the database.}
\label{fig:reaction-object-relationships}
\end{figure*}

DianShi-RxnDB centers its reaction data on the \dsReactionInstance{}, connecting source patent \dsReferences{} with chemical \dsSubstances{} and further organizing or associating reaction instances through \dsReactionGroups{} and \dsReactionTemplates{}.
This subsection describes these five core object types and their relationships; the conceptual relationships among \dsReactionInstance{}, \dsReactionGroup{}, and \dsReactionTemplate{} are shown in \Cref{fig:reaction-object-relationships}.

A source patent document is represented as a \dbobjectfirst{Reference}.
A \dsReference{} organizes the patent title, patent identifier, and other available document metadata.
One \dsReference{} can be linked to multiple \dsReactionInstances{}, and each \dsReactionInstance{} retains its source \dsReference{} together with the corresponding source-text location or region, supporting inspection of the patent context for procedures, participant roles, yields, and other recorded information.

A \dbobjectfirst{Substance} represents a chemical substance organized in the database and stores available information such as its name, molecular formula, molecular weight, canonical SMILES, and InChI.
It is connected to a \dsReactionInstance{} through a Reaction Participant relationship, and these relationships support browsing \dsSubstance{} records by associated patents, reaction instances, and participant roles.
A Reaction Participant records the role of a \dsSubstance{} in a particular single-step reaction as Reactant, Product, Reagent, Solvent, or Catalyst; the same \dsSubstance{} can take different roles in different \dsReactionInstances{}.

A \dbobjectfirst{Reaction Group} organizes \dsReactionInstances{} by a normalized reactant--product identity combination.
The same normalized reactant--product identity combination may be reported multiple times in different patents or under different experimental conditions, and these specific single-step reaction records are assigned to the same \dsReactionGroup{} for comparison.
A \dsReactionGroup{} does not merge these records into a single composite record; the reagents, catalysts, solvents, experimental conditions, yields, experimental procedures, and provenance information of each instance remain stored in the corresponding \dsReactionInstance{}.

A \dbobjectfirst{Reaction Template} represents a reaction transformation using SMARTS, provides a more abstract representation than a specific reactant--product combination, and is linked to corresponding \dsReactionInstances{} through database relationships.
The database maintains separate template collections generated using LocalRetro~\citep{chen2021localretro} and RDChiral~\citep{coley2019rdchiral}.

Definitions and counting conventions for the core objects are provided in \Cref{app:core-data-objects,app:patent-processing-counts}, while their principal fields are listed in \Cref{app:reference-fields,app:reaction-instance-fields,app:substance-fields}.

\subsection{Database scale and counting conventions}
\label{sec:database-scale}

\begin{table}[t]
\centering
\small
\caption{Scale of core DianShi-RxnDB data objects.}
\label{tab:database-object-scale}
\begin{tabular}{p{3.7cm}rp{7.2cm}}
\toprule
Database object &
Count &
Counting convention \\
\midrule
Source patent documents (\dsReferences{}) &
608,309 &
USPTO or EPO patent documents linked to at least one retained \dsReactionInstance{} \\
\dsSubstances{} &
6,261,797 &
\dsSubstance{} records organized by chemical identity in the database \\
\dsReactionInstances{} &
23,999,236 &
Specific reported experiments extracted from patents and retained in the database \\
\dsReactionGroups{} &
6,580,963 &
Groups formed from normalized reactant--product identity combinations \\
LocalRetro \dsReactionTemplates{} &
1,190,000 &
Deduplicated template collection generated using LocalRetro \\
RDChiral \dsReactionTemplates{} &
1,860,000 &
Deduplicated template collection generated using RDChiral \\
\bottomrule
\end{tabular}
\end{table}

\begin{table}[t]
\centering
\small
\caption{Automated qualification status of \dsReactionInstances{}.}
\label{tab:qualification-status}
\begin{tabular}{p{7.0cm}rr}
\toprule
Automated qualification status &
Count &
Share of all \dsReactionInstances{} \\
\midrule
Qualified instances &
14,808,205 &
61.7\% \\
Not qualified instances &
9,191,031 &
38.3\% \\
Total &
23,999,236 &
100\% \\
\bottomrule
\end{tabular}
\end{table}

The scale of the core database objects and the automated qualification results for the \dsReactionInstance{} population are reported in \Cref{tab:database-object-scale} and \Cref{tab:qualification-status}, respectively.

The approximately 24 million \dsReactionInstances{} represent the complete collection retained in the database, while the approximately 14.8 million qualified instances represent those that pass the automated qualification assessment.
The assessment marks whether an instance passes implemented checks based on atom mapping, atom conservation, and other rules.
Instances that do not pass the assessment nevertheless retain their extracted experimental information and provenance relationships.

The manual quality evaluation in \Cref{sec:manual-quality-evaluation} uses qualified instances as its target population and samples records only from that population.
Its results do not apply to \dsReactionInstances{} that did not pass the automated qualification assessment.

\subsection{Manual field-level quality evaluation of qualified \dsReactionInstances{}}
\label{sec:manual-quality-evaluation}

To evaluate the quality of core reaction fields in qualified instances produced by the fully automated pipeline, we treated the 14,808,205 \dsReactionInstances{} that passed the automated qualification assessment as the target population and randomly sampled 1,300 records from this population.
The evaluation covered five fields: Yield, Reactant, Reagent, Catalyst, and Solvent.
The evaluation was conducted by reviewers with professional backgrounds in organic chemistry, who independently compared each structured field with the corresponding experimental content in the source patent.
Disagreements were resolved through re-evaluation and adjudication according to the procedure described in \Cref{app:annotation-review}.

Each sampled record contributed one judgment for each of the five evaluated fields, yielding 6,500 field-level judgments.
Of these judgments, 6,042 were marked correct, corresponding to a micro-averaged field-level accuracy of 92.95\% across the five fields.
This result is used to estimate the quality of the five evaluated fields in the qualified-instance population.
\Cref{tab:manual-quality-results} reports the field-specific results.

Further details on the evaluation population, review procedure, and statistical definitions are provided in \Cref{app:quality-evaluation-protocol}.

\begin{table}[t]
\centering
\small
\caption{Manual quality-evaluation results for five reaction fields in qualified \dsReactionInstances{}.}
\label{tab:manual-quality-results}
\begin{tabular}{lrrr}
\toprule
Field &
Correct judgments &
Evaluated records &
Field-level accuracy \\
\midrule
Yield &
1,233 &
1,300 &
94.85\% \\
Reactant &
1,226 &
1,300 &
94.31\% \\
Reagent &
1,100 &
1,300 &
84.62\% \\
Catalyst &
1,265 &
1,300 &
97.31\% \\
Solvent &
1,218 &
1,300 &
93.69\% \\
Micro-average across five fields &
6,042 &
6,500 &
92.95\% \\
\bottomrule
\end{tabular}
\end{table}

Among the five evaluated fields, catalyst had the highest observed field-level accuracy at 97.31\%, whereas reagent had the lowest at 84.62\%.
Yield, reactant, and solvent ranged from 93.69\% to 94.85\%.
The boundary between reagents and other participant roles depends strongly on the specific reaction context, which may be one reason that reagent classification is more prone to error.

The reviewed errors fell into four recurring categories:

\begin{enumerate}
\item \textbf{Workup and purification information.} Materials used during drying, quenching, washing, or purification were sometimes recorded as reagents or solvents.

\item \textbf{Participant duplication and role assignment.} A material mentioned repeatedly in an experimental description could produce duplicate records or be assigned incorrectly among reactant, reagent, solvent, and catalyst roles.

\item \textbf{Compact expressions and cross-paragraph extraction.} Solvents in parenthetical or concentration expressions could be omitted, and yields reported in later paragraphs could be missed.

\item \textbf{Contextual references and reaction-step boundaries.} Information referenced from general procedures, other examples, or intermediate preparations could be incompletely resolved, and multiple reaction steps could be combined into one record.
\end{enumerate}

These observations identify directions for further data-quality improvement, including the treatment of workup information, participant-role assignment, cross-paragraph extraction, and multi-step experimental descriptions.

\subsection{External comparison with the Pistachio Reaction Dataset}
\label{sec:pistachio-comparison}

We compared DianShi-RxnDB with the Pistachio Reaction Dataset along three dimensions: reported reaction-record scale and matched-sample deduplication, representation granularity, and field-level exact agreement against source-grounded references.
Pistachio is a commercial dataset that is updated periodically, and the vendor's website documented a 2026Q2 release at the time of writing.
The record-level Pistachio data available to us were from the 2025Q2 release.
The record counts in the matched sample, the comparison of field-level exact agreement against source-grounded references, and the inspected shared-reaction example used for the representation comparison all refer to that release.
Pistachio was treated as an external comparator rather than as ground truth, and the field-level comparison was anchored to the corresponding patent source; detailed protocols and normalization rules are provided in \Cref{app:pistachio-comparison}.

\subsubsection{Scale and deduplication}
\label{sec:pistachio-scale}

At the reported-dataset level, DianShi-RxnDB contains approximately 24 million \dsReactionInstances{}, compared with approximately 21 million reported Pistachio reaction records~\citep{pistachio2026,mayfield2017pistachio,mayfield2021pistachio}.
These resource-level counts provide broad scale context but are not strictly commensurate because the two resources may use different record definitions and counting conventions.
Because the Pistachio release examined here provides separate text and image channels, the same patent example may be represented in both channels.
Under a common reactant--product-based deduplication rule, analysis of the matched sample of 100 US patents reduced DianShi-RxnDB from 4,222 to 4,093 retained records and Pistachio from 3,630 to 2,992 retained records, as shown in \Cref{tab:pistachio-deduplication}.
Thus, DianShi-RxnDB contained 1.368 times as many records retained after deduplication in this matched sample.
This ratio is descriptive of the matched sample and is not extrapolated to the complete corpora; the detailed deduplication key and channel-overlap counts are provided in Appendix~\ref{app:pistachio-comparison}.

\begin{table}[t]
\centering
\small
\caption{Reaction-record counts before and after deduplication in the matched sample of 100 US patents.}
\label{tab:pistachio-deduplication}
\begin{tabular}{lrrr}
\toprule
Dataset & Raw records & Records retained after deduplication & Duplicates removed \\
\midrule
DianShi-RxnDB & 4,222 & 4,093 & 129 \\
Pistachio, text + image & 3,630 & 2,992 & 638 \\
\bottomrule
\end{tabular}
\\[2pt]
\parbox{0.9\linewidth}{\footnotesize
Pistachio duplicates comprised 115 text-internal duplicates, 89 image-internal duplicates, and 434 text--image overlaps.}
\end{table}

\subsubsection{Field richness and representation granularity}
\label{sec:pistachio-granularity}

The inspected shared-reaction example from patent US10975080 illustrates how the two datasets organize and expose reaction information differently.
In this example, DianShi-RxnDB explicitly separates selected participant, process, workup, provenance, and validation features that are not exposed as corresponding standalone fields or objects in the inspected Pistachio record, as summarized in \Cref{tab:pistachio-representation}.
\begin{table}[t]
\centering
\footnotesize
\setlength{\tabcolsep}{3pt}
\renewcommand{\arraystretch}{1.08}
\caption{Selected differentiating representation features observed for the shared reaction \texttt{Intermediate 4} in patent US10975080. ``Yes'' indicates an explicitly represented feature; ``No'' indicates that no corresponding standalone field or object was exposed in the inspected record. This is not an exhaustive field inventory.}
\label{tab:pistachio-representation}
\begin{tabular}{p{5.0cm}cc}
\toprule
Selected differentiating feature & DianShi-RxnDB & Inspected Pistachio record \\
\midrule
Separate Reagent and Catalyst roles & \textbf{Yes} & \textbf{No}; broad \texttt{Agent} \\
Stage-level reaction object & \textbf{Yes} & \textbf{No} \\
Dedicated workup field & \textbf{Yes} & \textbf{No} \\
Explicit validation diagnostics & \textbf{Yes} & \textbf{No} \\
\bottomrule
\end{tabular}
\end{table}

Both records also contain product structures, quantities, temperature, time, yield information, and an ordered procedure, although these elements are organized differently.
A more detailed record-level comparison is provided in \Cref{tab:app-pistachio-representation}.

\subsubsection{Field-level exact agreement against source-grounded references}
\label{sec:pistachio-quality}

For the comparison of field-level exact agreement against source-grounded references, we identified 660 reaction pairs from 58 patents for which both records belonged to the same patent and had exactly matching normalized source paragraphs from the patent XML/SGM.
Source-paragraph normalization lowercased the text and removed markup, punctuation, whitespace, and line-break differences; this criterion establishes common source provenance but does not imply that either extraction is correct.
For Reactant, Reagent, Catalyst, Solvent, Product, and Yield, GPT-5.6-sol performed LLM-assisted semantic adjudication under the field definitions, while deterministic code performed normalization, field-level agreement assessment, counting, and percentage calculation.
Full details of reference construction, role harmonization, adjudication, and scoring are provided in \Cref{app:pistachio-quality}.

As shown in \Cref{tab:pistachio-field-comparison}, DianShi-RxnDB showed numerically higher field-level exact agreement against source-grounded references than Pistachio for all six evaluated fields across the 660 source-paragraph-matched reaction pairs.
The differences were relatively small for Reactant, Solvent, and Product, whereas the largest difference occurred for Yield.
This Yield difference is affected in part by representation policy: DianShi-RxnDB generally records explicit percentages from source text, while Pistachio may provide values inferred from mass and stoichiometry when no percentage is explicitly stated.
\begin{table}[t]
\centering
\small
\caption{Field-level exact agreement against source-grounded references in 660 source-paragraph-matched reaction pairs.}
\label{tab:pistachio-field-comparison}
\begin{tabular}{lccc}
\toprule
Field & Pistachio & DianShi-RxnDB & Gap \\
\midrule
Reactant & 623/660 (94.39\%) & 643/660 (97.42\%) & +3.03\% \\
Reagent & 555/660 (84.09\%) & 576/660 (87.27\%) & +3.18\% \\
Catalyst & 553/660 (83.79\%) & 569/660 (86.21\%) & +2.42\% \\
Solvent & 615/660 (93.18\%) & 629/660 (95.30\%) & +2.12\% \\
Product & 604/660 (91.52\%) & 609/660 (92.27\%) & +0.75\% \\
Yield & 535/660 (81.06\%) & 656/660 (99.39\%) & +18.33\% \\
\bottomrule
\end{tabular}
\end{table}

Together, the results of the external matched comparison indicate a larger reaction-record count after deduplication, finer representation granularity in the inspected shared-reaction example, and numerically higher field-level exact agreement against source-grounded references across the six evaluated fields.

\section{Web research workbench}
\label{sec:web-workbench}

The preceding section described the data foundation, core objects, and quality evaluation of DianShi-RxnDB.
On this foundation, the Web research workbench and MCP service form two complementary interfaces for researchers and AI agents, respectively.
This section first describes the interactive search, linked exploration, and source-verification capabilities of the Web research workbench; the MCP service is described in the following section.

\subsection{Search entry points and result organization}
\label{sec:web-search-results}

Building on the core-object system described in \Cref{sec:database-objects}, the Web research workbench provides three search entry points for substances, reactions, and patent \dsReferences{} and organizes query conditions and results by object type.

\dsSubstance{} search supports queries by name or molecular structure.
Researchers can enter a structure representation directly or draw a chemical structure using the structure editor.
Results are organized as \dsSubstance{} records, display basic information such as molecular structure, name, molecular formula, and molecular weight, and support filtering related records by the participant role of the \dsSubstance{} in a reaction.

Reaction search supports queries using Reaction SMILES or Reaction SMARTS.
Results are organized at two levels of granularity: a \dsReactionInstance{} presents a specific single-step reaction record, while a \dsReactionGroup{} aggregates instances sharing the same reactant--product combination; the experimental information of each instance remains separate within the group.

\dsReference{} search supports queries by patent identifier or by text over titles, abstracts, and keywords.
Results are organized as \dsReference{} records and display basic information such as patent title, identifier, date, and abstract.

\subsection{Linked exploration and instance comparison}
\label{sec:web-linked-exploration}

The Web research workbench presents the object relationships described in \Cref{sec:database-objects} as continuous browsing paths.
Researchers can navigate from a \dsSubstance{} or \dsReference{} to related \dsReactionInstances{}, or from a specific \dsReactionInstance{} to its participants, \dsReactionGroup{}, and source \dsReference{}, without issuing separate independent queries.

Within a \dsReactionGroup{}, researchers can expand multiple \dsReactionInstances{} sharing the same reactant--product combination and compare their separately recorded participant roles, experimental conditions, yields, procedures, and source information.
The workbench does not merge these instances into a composite record; the next subsection describes how to return from a specific instance to its source patent for source-text verification.

\subsection{Provenance linkage and source-text verification}
\label{sec:web-source-verification}

\begin{figure*}[t]
\centering
\includegraphics[width=\textwidth]{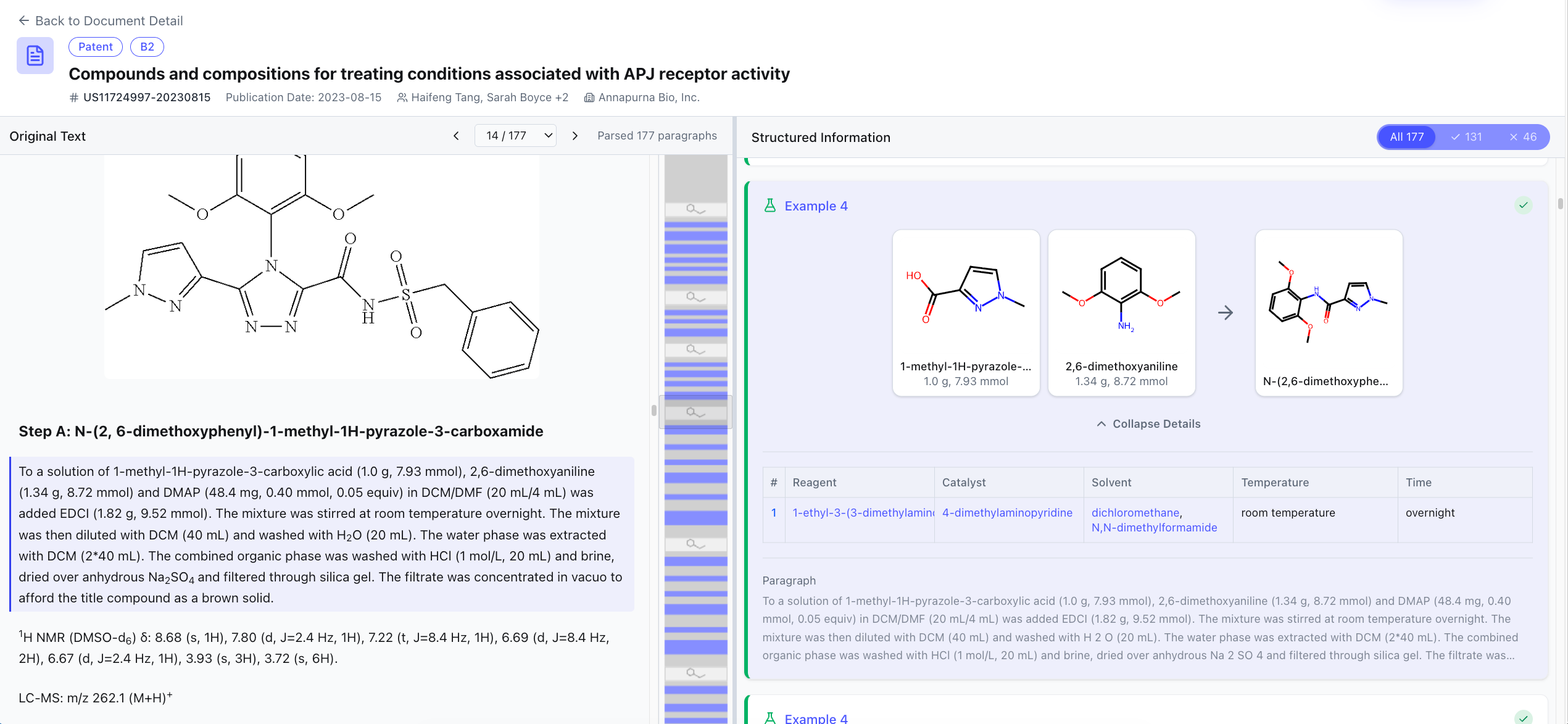}
\caption{Structured \dsReactionInstance{} and source-patent comparison in the Web research workbench.
The left panel shows experimental content from the source patent, while the right panel shows structured reaction information for the corresponding \dsReactionInstance{}, including reaction structures, participants, amounts, and experimental conditions.
For records with available source content and location information, users can locate and compare the relevant source-patent paragraph or text region in the same interface to verify experimental procedures, participant roles, experimental conditions, and yields.}
\label{fig:web-source-linked-inspection}
\end{figure*}

Each \dsReactionInstance{} is linked to its source \dsReference{} and the corresponding paragraph or region in the source text.
The Web research workbench supports viewing a structured record and its source-patent context in the same interface and navigating from a \dsReactionInstance{} precisely to the relevant source-patent paragraph or text region, as shown in \Cref{fig:web-source-linked-inspection}.
Researchers can use this linkage to verify experimental procedures, participant roles, experimental conditions, yields, and other information and to identify possible omissions or ambiguities in the structured record.

\section{MCP service for AI agents}
\label{sec:mcp-service}

\subsection{Role of the MCP service}
\label{sec:mcp-role}

DianShi-RxnDB provides AI agents with a structured chemical-information retrieval interface through the Model Context Protocol (MCP).
On the same data foundation, the Web research workbench supports interactive retrieval, instance comparison, and source verification by researchers, while the MCP service organizes query and retrieval capabilities for substances, reactions, and patent documents as composable structured tools, supporting AI agents in invoking relevant database records and conducting multi-step retrieval.

\subsection{MCP tool capabilities}
\label{sec:mcp-capabilities}

The MCP service primarily provides four categories of structured retrieval capabilities:

\begin{description}
\item[\dsSubstance{} retrieval.]
The service retrieves \dsSubstance{} records by name, database identifier, or chemical representation and supports structure-based similarity and substructure queries.

\item[\dsReactionGroup{} retrieval.]
The service retrieves relevant \dsReactionGroups{} using a reaction representation or target product.

\item[\dsReactionInstance{} access.]
The service retrieves \dsReactionInstances{} associated with a \dsReactionGroup{}, together with recorded participant roles, experimental conditions, yields, and source information.

\item[Source \dsReference{} retrieval.]
The service retrieves a \dsReference{} by database identifier or finds relevant patent documents through a text query.
\end{description}

\subsection{Multi-step retrieval and source linkage}
\label{sec:mcp-multistep-retrieval}

MCP tools return query results as structured fields.
Database object identifiers and chemical representations in one response can be used as inputs to subsequent calls, forming a multi-step retrieval path that includes candidate retrieval, target identification, instance access, and source lookup.
A reaction-precedent search can, for example, proceed as follows:

\begin{quote}
User question
$\rightarrow$ retrieve candidate \dsReactionGroups{} by target product
$\rightarrow$ identify the target \dsReactionGroup{} from the reactant--product combination
$\rightarrow$ obtain associated \dsReactionInstances{} and their experimental information
$\rightarrow$ query source \dsReferences{}
$\rightarrow$ organize the structured results and their sources.
\end{quote}

Results retrieved by target product may contain multiple reactant combinations.
An agent must use the task constraints to identify the target \dsReactionGroup{} before obtaining its associated \dsReactionInstances{}.
The case study in \Cref{sec:use-cases} demonstrates this stepwise filtering process.

MCP returns source \dsReferences{} so that structured results remain associated with the corresponding patent documents, while interactive positioning to a source-text paragraph or region and contextual reading are provided by the Web research workbench.
Researchers can use chemical representations and source information returned by MCP to search for the corresponding records in the Web workbench and further verify the source content.

\section{Reaction-precedent retrieval case study with Web and MCP}
\label{sec:use-cases}

This chapter uses the classical synthesis of aspirin as an illustrative case to present two usage paths.
In the first, a researcher uses the Web research workbench directly to retrieve, compare, and inspect reaction instances and their sources; in the second, an AI agent retrieves reaction instances and source information through MCP.
Both paths concern the same target reaction and the same set of database objects and illustrate the complementary roles of Web and MCP in information discovery, instance comparison, and source inspection.

\subsection{Retrieving reaction instances through the Web workbench}
\label{sec:web-aspirin-case}

This case uses aspirin, or acetylsalicylic acid, as the target compound.
Its SMILES is:

\begin{quote}
\texttt{CC(=O)Oc1ccccc1C(=O)O}
\end{quote}

\begin{figure*}[t]
\centering
\includegraphics[width=\textwidth]{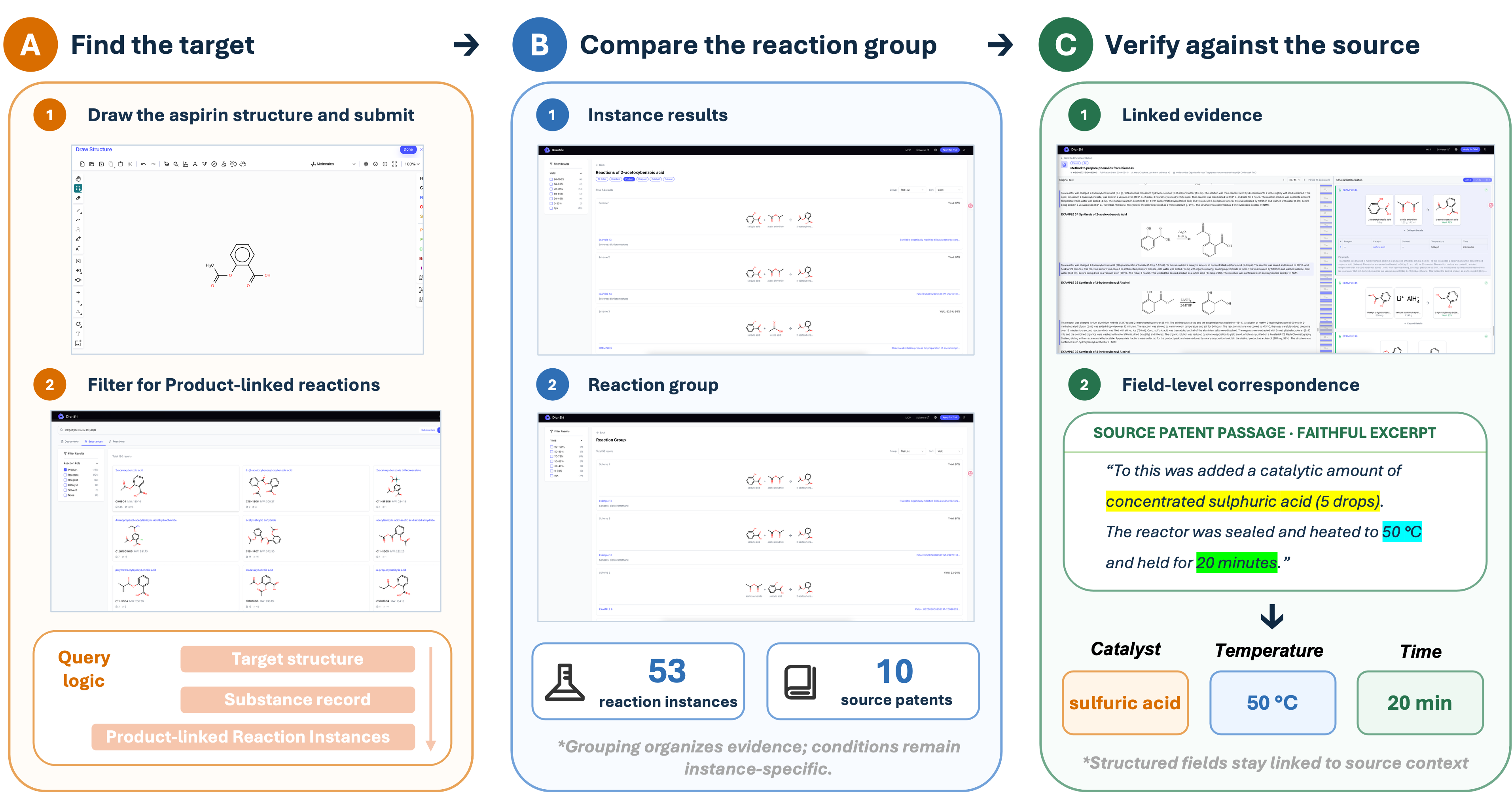}
\caption{Web use case from an aspirin target structure to patent reaction precedents and source verification.
(A) The user enters or draws the aspirin structure, matches the corresponding \dsSubstance{}, restricts its participant role to Product, and enters the related reaction instances.
(B) Among the 84 \dsReactionInstances{} associated with aspirin as a Product, the user locates the target \dsReactionGroup{} in which salicylic acid and acetic anhydride form aspirin.
This group contains 53 \dsReactionInstances{} from 10 source patents, with each instance's experimental conditions, yield, and source information retained separately for comparison.
(C) The user opens the source patent associated with Instance A and verifies the catalyst sulfuric acid, temperature of 50\,$^\circ$C, and time of 20 minutes in the structured record.}
\label{fig:web-aspirin-case}
\end{figure*}

The researcher enters or draws this structure in the Web research workbench, locates the corresponding \dsSubstance{}, and restricts the participant role of aspirin to Product.
The retrieval and filtering process is illustrated in panels A and B of \Cref{fig:web-aspirin-case}.
In this case, the result page displays 84 related \dsReactionInstances{} in which aspirin is a Product.
These records contain different reactant combinations and do not all belong to the same synthetic route.
The researcher then selects the classical acetylation in which salicylic acid and acetic anhydride are the reactants:

\begin{quote}
Salicylic acid $+$ acetic anhydride $\rightarrow$ aspirin.
\end{quote}

The target \dsReactionGroup{} contains 53 \dsReactionInstances{} in this query.

To compare the information recorded for single-step reaction records within the same \dsReactionGroup{}, we selected Instances A, B, and C from different source patents.
Instance B reports the addition of swellable organically modified silica (SOMS).
\Cref{tab:web-aspirin-instances} primarily reports the structured Web records.
Temperature, time, and percentage ranges are formatted consistently for readability, but fields not recorded in the structured records are not silently completed from chemical knowledge or the source patent.

\begin{table*}[t]
\centering
\small
\setlength{\tabcolsep}{2.3pt}
\caption{Representative patent reaction instances for the formation of aspirin from salicylic acid and acetic anhydride.}
\label{tab:web-aspirin-instances}
\begin{tabular}{p{1.1cm}p{2.4cm}p{2.4cm}p{1.3cm}p{1.6cm}p{1.9cm}p{3.0cm}}
\toprule
\shortstack[l]{Instance} &
\shortstack[l]{Reagent /\\catalyst} &
\shortstack[l]{Solvent} &
\shortstack[l]{Temp.} &
\shortstack[l]{Time} &
\shortstack[l]{Yield} &
\shortstack[l]{Source patent} \\
\midrule
A &
Sulfuric acid &
Not recorded &
50\,$^\circ$C &
20 min &
70\% &
\shortstack[l]{\scriptsize US10407376B2\\\scriptsize \citep{crockatt2019phenolics}} \\
B &
SOMS &
Dichloromethane &
40\,$^\circ$C &
Not rec. &
97\% &
\shortstack[l]{\scriptsize US20220008887A1\\\scriptsize \citep{shaw2022soms}} \\
C &
Sulfated zirconia &
Not recorded &
120\,$^\circ$C &
30 min &
92--95\% &
\shortstack[l]{\scriptsize US20090082592A1\\\scriptsize \citep{jasra2009aspirin}} \\
\bottomrule
\end{tabular}
\end{table*}

To check the structured result, we opened the source patent associated with Instance A, using the linked view shown in panel C of \Cref{fig:web-aspirin-case}.
The source text reports the addition of the quoted phrase \textit{a catalytic amount of concentrated sulphuric acid (5 drops)} and heating the reaction mixture at 50\,$^\circ$C for 20 minutes, consistent with the catalyst, temperature, and time in the structured Web record~\citep{crockatt2019phenolics}.
The patent also describes cooling, precipitation by adding ice water, filtration, washing, and vacuum drying.

\subsection{Retrieving reaction instances through MCP}
\label{sec:mcp-aspirin-case}

To illustrate the MCP retrieval process, this subsection uses the same target reaction as \Cref{sec:web-aspirin-case}.
After connecting the DianShi MCP service to a compatible client, we submitted the following task to an AI agent:

\begin{quote}
Find reaction records in DianShi-RxnDB in which salicylic acid and acetic anhydride are reactants and aspirin (\texttt{CC(=O)Oc1ccccc1C(=O)O}) is the Product.
Summarize the reagent or catalyst, solvent, temperature, reaction time, yield, and source patent for selected reported instances, and provide database object identifiers that can support subsequent verification.
Summarize only information returned by the database; explicitly mark missing fields, do not fill them from chemical knowledge, and do not present reported conditions as recommended conditions.
\end{quote}

The actual tool calls formed the following retrieval path:

\begin{quote}
Natural-language task
$\rightarrow$ retrieve 11 candidate \dsReactionGroups{} by aspirin product structure
$\rightarrow$ identify 1 target \dsReactionGroup{} from the salicylic-acid--acetic-anhydride reactant combination
$\rightarrow$ confirm that the target group contains 53 \dsReactionInstances{}
$\rightarrow$ return 20 of those instances in this condition query
$\rightarrow$ query the corresponding \dsReferences{} and organize the answer.
\end{quote}

Based on the results returned in this call, the agent organized the target reaction, experimental conditions, yields, source patents, and database object identifiers.
Fields not returned by the tools were explicitly marked as missing.

\begin{figure*}[t]
\centering
\includegraphics[width=\textwidth]{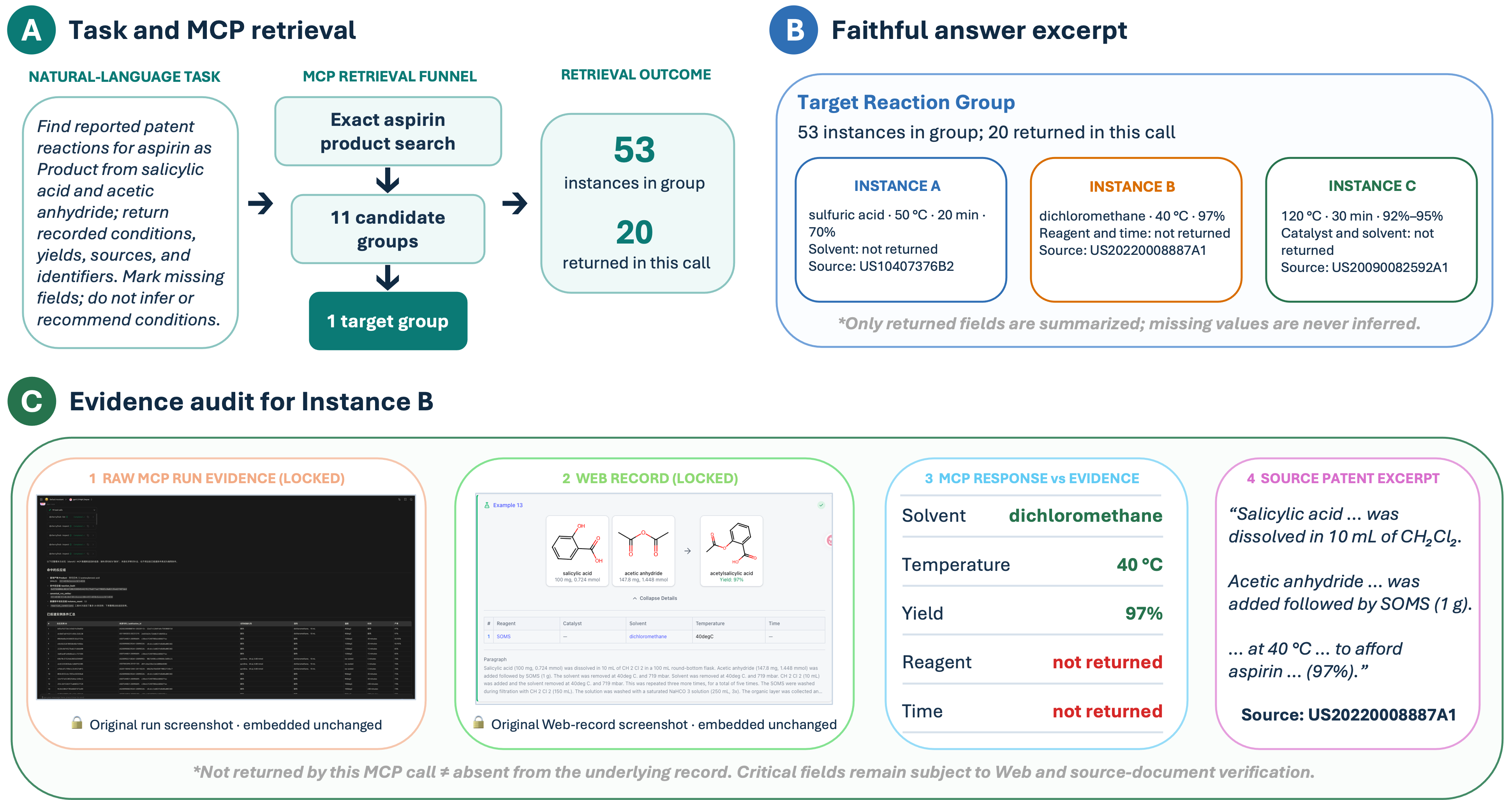}
\caption{From a natural-language task to multi-step MCP retrieval and example source inspection.
(A) The AI agent calls the DianShi MCP according to the natural-language task: it retrieves 11 candidate \dsReactionGroups{} by the aspirin product structure, then locates 1 target \dsReactionGroup{} from the salicylic-acid--acetic-anhydride reactant combination.
The group contains 53 \dsReactionInstances{}, of which 20 are returned by the condition query in this call.
(B) Based on the MCP results returned in this call, the agent organizes the target \dsReactionGroup{} and selected representative \dsReactionInstances{} with their experimental conditions, yields, and source information.
(C) Using Instance B as an example, the agent uses the reaction and source information returned by MCP to locate the corresponding record in the Web research workbench and inspect the source patent for SOMS, dichloromethane, 40\,$^\circ$C, and 97\% as reported experimental information.}
\label{fig:mcp-aspirin-case}
\end{figure*}

Chemical representations and source information returned by MCP provide clues for researchers to continue inspecting the corresponding records.
Researchers can use them to locate related records in the Web research workbench and view their source-patent context.
The process from the natural-language task and MCP tool calls to structured results and source-patent comparison is shown in \Cref{fig:mcp-aspirin-case}.

This case shows that DianShi-RxnDB can provide researchers and AI agents with two complementary reaction-precedent retrieval paths over the same data foundation.
MCP supports AI agents in conducting structured, multi-step retrieval and organizing reaction records with source information, while the Web research workbench supports researchers in further comparing single-step reaction records, viewing source-patent context, and verifying key information.

\section{Limitations, availability, and responsible use}
\label{sec:limitations-responsible-use}

\subsection{Data and use boundaries}
\label{sec:data-use-limitations}

The data foundation described in this report consists of organic synthesis patent information from the USPTO and EPO and primarily covers patents published from 1976 through 2025.
Its coverage is also affected by patent-record consolidation rules and the database update cycle.
Journal articles and unpublished laboratory knowledge are outside the current data scope.

The manual quality-evaluation results in \Cref{sec:manual-quality-evaluation} are used to estimate the quality of the Yield, Reactant, Reagent, Catalyst, and Solvent fields in the qualified-instance population.
They do not apply to \dsReactionInstances{} that did not pass the automated qualification assessment and do not represent the overall correctness of an individual \dsReactionInstance{}, accuracy for unevaluated fields, or database recall.
Automated extraction can still produce field omissions, duplicate records, role errors, and step-boundary errors; users can verify critical records against the source patents.

\subsection{AI agents and responsible use}
\label{sec:agent-responsible-use}

Structured records, as well as an agent's organization and interpretation of retrieval results, may contain omissions or errors.
Provenance linkage provides a path back to the patent document for verification but does not replace the judgment of qualified professionals working from the complete context.

For uses involving experimental safety, clinical or regulatory matters, patent assessment, or production-condition selection, the relevant information should be further reviewed by professionals with the appropriate expertise.

\subsection{Access and licensing}
\label{sec:availability}

DianShi-RxnDB provides online access through the Web research workbench and MCP service.
Web and MCP access is available free of charge for non-commercial use, while commercial use requires a separate license; specific access points, connection procedures, and applicable terms are provided in the platform's official documentation.

\section{Conclusion and outlook}
\label{sec:conclusion}

This report introduces DianShi-RxnDB, a large-scale, fine-grained organic reaction data platform for organic chemistry researchers and AI agents.
The platform uses a fully automated information-extraction and normalization pipeline covering patent text, images, and reaction schemes to organize experimental knowledge from organic synthesis patents from the USPTO and EPO into structured \dsReactionInstances{}.
The database contains approximately 24 million \dsReactionInstances{}, of which approximately 14.8 million (61.7\%) pass the automated qualification assessment; each instance can organize participant roles, quantities, experimental conditions, yields, and experimental procedures, while structured relationships are established among \dsSubstances{}, \dsReactionInstances{}, \dsReactionGroups{}, and patent \dsReferences{}.
We randomly sampled 1,300 records from the qualified-instance population for manual quality evaluation, obtaining a micro-averaged accuracy of 92.95\% across the Yield, Reactant, Reagent, Catalyst, and Solvent fields.
The matched comparison with the Pistachio Reaction Dataset further showed that DianShi-RxnDB retained more reaction records after deduplication, exhibited finer-grained information organization in the inspected representative record, and achieved higher field-level exact agreement against source-grounded references across all six evaluated fields.

The Web research workbench and MCP service operate on the same organic reaction data foundation.
They support interactive retrieval, instance comparison, and source inspection by researchers, and structured tool calls and multi-step retrieval by AI agents, respectively.
Provenance linkage allows researchers to return from structured records to source-patent text when needed and inspect critical information.
This dual-interface design demonstrates how a shared, provenance-linked data foundation can support both researcher-facing investigation and agent-based retrieval.

Future work will expand the sources of organic reaction data, continue to improve data quality, enhance retrieval and source-verification capabilities for researchers and AI agents, and explore applications of DianShi-RxnDB in a broader range of organic chemistry research scenarios; the next technical report will focus on applying this data foundation to retrosynthesis tasks and presenting initial research findings.

\section{Acknowledgments}
\label{sec:acknowledgments}

This project was supported by Shanghai Artificial Intelligence Laboratory.

\clearpage
\bibliographystyle{plainnat}
\setcitestyle{numbers}
\bibliography{paper}

\begin{thebibliography}{37}
\providecommand{\natexlab}[1]{#1}
\providecommand{\url}[1]{\texttt{#1}}
\expandafter\ifx\csname urlstyle\endcsname\relax
  \providecommand{\doi}[1]{doi: #1}\else
  \providecommand{\doi}{doi: \begingroup \urlstyle{rm}\Url}\fi

\bibitem[{CAS}(2026{\natexlab{a}})]{casReactions}
{CAS}.
\newblock {CAS Reactions}, 2026{\natexlab{a}}.
\newblock URL \url{https://www.cas.org/cas-data/cas-reactions}.
\newblock Accessed 2026-08-22.

\bibitem[{CAS}(2026{\natexlab{b}})]{casSciFinder}
{CAS}.
\newblock {CAS SciFinder}, 2026{\natexlab{b}}.
\newblock URL \url{https://www.cas.org/solutions/cas-scifinder-discovery-platform/cas-scifinder}.
\newblock Accessed 2026-08-22.

\bibitem[Chen and Jung(2021)]{chen2021localretro}
Shuan Chen and Yousung Jung.
\newblock Deep retrosynthetic reaction prediction using local reactivity and global attention.
\newblock \emph{JACS Au}, 1\penalty0 (10):\penalty0 1612--1620, 2021.
\newblock \doi{10.1021/jacsau.1c00246}.

\bibitem[Coley et~al.(2017)Coley, Barzilay, Jaakkola, Green, and Jensen]{coley2017outcomes}
Connor~W. Coley, Regina Barzilay, Tommi~S. Jaakkola, William~H. Green, and Klavs~F. Jensen.
\newblock Prediction of organic reaction outcomes using machine learning.
\newblock \emph{ACS Central Science}, 3\penalty0 (5):\penalty0 434--443, 2017.
\newblock \doi{10.1021/acscentsci.7b00064}.

\bibitem[Coley et~al.(2019)Coley, Green, and Jensen]{coley2019rdchiral}
Connor~W. Coley, William~H. Green, and Klavs~F. Jensen.
\newblock {RDChiral}: An {RDKit} wrapper for handling stereochemistry in retrosynthetic template extraction and application.
\newblock \emph{Journal of Chemical Information and Modeling}, 59\penalty0 (6):\penalty0 2529--2537, 2019.
\newblock \doi{10.1021/acs.jcim.9b00286}.

\bibitem[Crockatt et~al.(2019)Crockatt, Urbanus, Konst, and de~Koning]{crockatt2019phenolics}
Marc Crockatt, Jan~Harm Urbanus, Paul~Mathijs Konst, and Martijn~Constantijn de~Koning.
\newblock Method to prepare phenolics from biomass, 2019.
\newblock URL \url{https://patents.google.com/patent/US10407376B2/en}.
\newblock U.S. Patent 10,407,376 B2.

\bibitem[{DeepLink}(2026)]{deeplink2026}
{DeepLink}.
\newblock {DeepLink}, 2026.
\newblock URL \url{https://deeplink.org.cn/}.

\bibitem[{DeepSeek-AI}(2025)]{deepseek2025v30324}
{DeepSeek-AI}.
\newblock {DeepSeek-V3-0324}, 2025.
\newblock URL \url{https://huggingface.co/deepseek-ai/DeepSeek-V3-0324}.
\newblock Hugging Face model card.

\bibitem[{Elsevier}(2026{\natexlab{a}})]{reaxysAcademic}
{Elsevier}.
\newblock Reaxys for academic research, 2026{\natexlab{a}}.
\newblock URL \url{https://www.elsevier.com/products/reaxys/higher-education/academic-research}.
\newblock Accessed 2026-08-22.

\bibitem[{Elsevier}(2026{\natexlab{b}})]{reaxysProduct}
{Elsevier}.
\newblock {Reaxys}: Chemistry data and {AI} to optimize small molecule discovery, 2026{\natexlab{b}}.
\newblock URL \url{https://www.elsevier.com/products/reaxys}.
\newblock Accessed 2026-08-22.

\bibitem[Fan et~al.(2024)Fan, Qian, Wang, Wang, Coley, and Barzilay]{fan2024openchemie}
Vincent Fan, Yujie Qian, Alex Wang, Amber Wang, Connor~W. Coley, and Regina Barzilay.
\newblock {OpenChemIE}: An information extraction toolkit for chemistry literature.
\newblock \emph{Journal of Chemical Information and Modeling}, 64\penalty0 (14):\penalty0 5521--5534, 2024.
\newblock \doi{10.1021/acs.jcim.4c00572}.

\bibitem[Gao et~al.(2018)Gao, Struble, Coley, Wang, Green, and Jensen]{gao2018conditions}
Hanyu Gao, Thomas~J. Struble, Connor~W. Coley, Yuran Wang, William~H. Green, and Klavs~F. Jensen.
\newblock Using machine learning to predict suitable conditions for organic reactions.
\newblock \emph{ACS Central Science}, 4\penalty0 (11):\penalty0 1465--1476, 2018.
\newblock \doi{10.1021/acscentsci.8b00357}.

\bibitem[Gong and collaborators(2024)]{gong2024usptollm}
Shukai Gong and collaborators.
\newblock {USPTO\_LLM}, 2024.
\newblock URL \url{https://github.com/GONGSHUKAI/USPTO_LLM}.

\bibitem[Guo et~al.(2022)Guo, Ibanez-Lopez, Gao, Quach, Coley, Jensen, and Barzilay]{guo2022reactionextraction}
Jiang Guo, Alvaro~S. Ibanez-Lopez, Hanyu Gao, Victor Quach, Connor~W. Coley, Klavs~F. Jensen, and Regina Barzilay.
\newblock Automated chemical reaction extraction from scientific literature.
\newblock \emph{Journal of Chemical Information and Modeling}, 62\penalty0 (9):\penalty0 2035--2045, 2022.
\newblock \doi{10.1021/acs.jcim.1c00284}.

\bibitem[Hasic and Ishida(2026)]{hasic2026consolidation}
Haris Hasic and Tetsuya Ishida.
\newblock The consolidation of open-source computer-assisted chemical synthesis data into a comprehensive database.
\newblock \emph{Journal of Cheminformatics}, 18:\penalty0 4, 2026.
\newblock \doi{10.1186/s13321-025-01130-0}.

\bibitem[Hawizy et~al.(2011)Hawizy, Jessop, Adams, and Murray-Rust]{hawizy2011chemicaltagger}
Lezan Hawizy, David~M. Jessop, Nico Adams, and Peter Murray-Rust.
\newblock {ChemicalTagger}: A tool for semantic text-mining in chemistry.
\newblock \emph{Journal of Cheminformatics}, 3:\penalty0 17, 2011.
\newblock \doi{10.1186/1758-2946-3-17}.

\bibitem[Jasra et~al.(2009)Jasra, Tyagi, and Mishra]{jasra2009aspirin}
Raksh~Vir Jasra, Beena Tyagi, and Manish~Kumar Mishra.
\newblock Green catalytic process for the synthesis of acetyl salicylic acid, 2009.
\newblock URL \url{https://patents.google.com/patent/US20090082592A1/en}.
\newblock U.S. Patent Application Publication US 2009/0082592 A1.

\bibitem[Jiang and collaborators(2021)]{jiang2021data}
Shu Jiang and collaborators.
\newblock Data for ``when {SMILES} smiles'', 2021.
\newblock URL \url{https://github.com/jshmjs45/data_for_chem}.

\bibitem[Jiang et~al.(2021)Jiang, Zhang, Zhao, Li, Yang, Lu, and Xia]{jiang2021smiles}
Shu Jiang, Zhuosheng Zhang, Hai Zhao, Jiangtong Li, Yang Yang, Bao-Liang Lu, and Ning Xia.
\newblock When {SMILES} smiles, practicality judgment and yield prediction of chemical reaction via deep chemical language processing.
\newblock \emph{IEEE Access}, 9:\penalty0 85071--85083, 2021.
\newblock \doi{10.1109/ACCESS.2021.3083838}.

\bibitem[Kearnes et~al.(2021)Kearnes, Maser, Wleklinski, Kast, Doyle, Dreher, Hawkins, Jensen, and Coley]{kearnes2021ord}
Steven~M. Kearnes, Michael~R. Maser, Michael Wleklinski, Anton Kast, Abigail~G. Doyle, Spencer~D. Dreher, Joel~M. Hawkins, Klavs~F. Jensen, and Connor~W. Coley.
\newblock The open reaction database.
\newblock \emph{Journal of the American Chemical Society}, 143\penalty0 (45):\penalty0 18820--18826, 2021.
\newblock \doi{10.1021/jacs.1c09820}.

\bibitem[Lowe(2017)]{lowe2017uspto}
Daniel Lowe.
\newblock Chemical reactions from {US} patents (1976--sep 2016), 2017.
\newblock URL \url{https://doi.org/10.6084/m9.figshare.5104873.v1}.

\bibitem[Lowe(2012)]{lowe2012thesis}
Daniel~Mark Lowe.
\newblock \emph{Extraction of Chemical Structures and Reactions from the Literature}.
\newblock PhD thesis, University of Cambridge, 2012.
\newblock URL \url{https://www.repository.cam.ac.uk/handle/1810/244727}.

\bibitem[Mayfield et~al.(2017)Mayfield, Lowe, and Sayle]{mayfield2017pistachio}
John Mayfield, Daniel Lowe, and Roger Sayle.
\newblock Pistachio: Search and faceting of large reaction databases, 2017.
\newblock URL \url{https://nextmovesoftware.com/blog/2017/12/11/pistachio-search-and-faceting-of-large-reaction-databases}.
\newblock ACS Fall 2017.

\bibitem[Mayfield et~al.(2021)Mayfield, Lowe, and Sayle]{mayfield2021pistachio}
John Mayfield, Daniel Lowe, and Roger Sayle.
\newblock Pistachio, 2021.
\newblock URL \url{https://nextmovesoftware.com/talks/Mayfield_Pistachio_NIHReactions_202105.pdf}.
\newblock NIH Virtual Workshop on Reaction Informatics.

\bibitem[{Model Context Protocol}(2026)]{mcp2026spec}
{Model Context Protocol}.
\newblock Model context protocol specification, 2026.
\newblock URL \url{https://modelcontextprotocol.io/specification/2026-07-28}.
\newblock Revision 2026-07-28.

\bibitem[{NextMove Software}(2026)]{pistachio2026}
{NextMove Software}.
\newblock Pistachio, 2026.
\newblock URL \url{https://www.nextmovesoftware.com/pistachio}.
\newblock Version 2026Q2; accessed 2026-08-22.

\bibitem[Schneider et~al.(2016)Schneider, Lowe, Sayle, Tarselli, and Landrum]{schneider2016patents}
Nadine Schneider, Daniel~M. Lowe, Roger~A. Sayle, Michael~A. Tarselli, and Gregory~A. Landrum.
\newblock Big data from pharmaceutical patents: A computational analysis of medicinal chemists' bread and butter.
\newblock \emph{Journal of Medicinal Chemistry}, 59\penalty0 (9):\penalty0 4385--4402, 2016.
\newblock \doi{10.1021/acs.jmedchem.6b00153}.

\bibitem[Schwaller et~al.(2019)Schwaller, Laino, Gaudin, Bolgar, Hunter, Bekas, and Lee]{schwaller2019molecular}
Philippe Schwaller, Teodoro Laino, Th{\'e}ophile Gaudin, Peter Bolgar, Christopher~A. Hunter, Costas Bekas, and Alpha~A. Lee.
\newblock Molecular transformer: A model for uncertainty-calibrated chemical reaction prediction.
\newblock \emph{ACS Central Science}, 5\penalty0 (9):\penalty0 1572--1583, 2019.
\newblock \doi{10.1021/acscentsci.9b00576}.

\bibitem[Schwaller et~al.(2021)Schwaller, Hoover, Reymond, Strobelt, and Laino]{schwaller2021rxnmapper}
Philippe Schwaller, Benjamin Hoover, Jean-Louis Reymond, Hendrik Strobelt, and Teodoro Laino.
\newblock Extraction of organic chemistry grammar from unsupervised learning of chemical reactions.
\newblock \emph{Science Advances}, 7\penalty0 (15):\penalty0 eabe4166, 2021.
\newblock \doi{10.1126/sciadv.abe4166}.

\bibitem[{Sciverse}(2026)]{sciverse2026scibase}
{Sciverse}.
\newblock {Sciverse}, 2026.
\newblock URL \url{https://sciverse.space/}.

\bibitem[Segler et~al.(2018)Segler, Preuss, and Waller]{segler2018planning}
Marwin H.~S. Segler, Mike Preuss, and Mark~P. Waller.
\newblock Planning chemical syntheses with deep neural networks and symbolic {AI}.
\newblock \emph{Nature}, 555\penalty0 (7698):\penalty0 604--610, 2018.
\newblock \doi{10.1038/nature25978}.

\bibitem[Shaw(2022)]{shaw2022soms}
Nicholas~N. Shaw.
\newblock Swellable organically modified silica as nanoreactors, 2022.
\newblock URL \url{https://patents.google.com/patent/US20220008887A1/en}.
\newblock U.S. Patent Application Publication US 2022/0008887 A1.

\bibitem[Thakkar et~al.(2020)Thakkar, Kogej, Reymond, Engkvist, and Bjerrum]{thakkar2020datasets}
Amol Thakkar, Thierry Kogej, Jean-Louis Reymond, Ola Engkvist, and Esben~Jannik Bjerrum.
\newblock Datasets and their influence on the development of computer assisted synthesis planning tools in the pharmaceutical domain.
\newblock \emph{Chemical Science}, 11\penalty0 (1):\penalty0 154--168, 2020.
\newblock \doi{10.1039/C9SC04944D}.

\bibitem[Vaucher et~al.(2020)Vaucher, Zipoli, Geluykens, Nair, Schwaller, and Laino]{vaucher2020actions}
Alain~C. Vaucher, Federico Zipoli, Joppe Geluykens, Vishnu~H. Nair, Philippe Schwaller, and Teodoro Laino.
\newblock Automated extraction of chemical synthesis actions from experimental procedures.
\newblock \emph{Nature Communications}, 11:\penalty0 3601, 2020.
\newblock \doi{10.1038/s41467-020-17266-6}.

\bibitem[Yang et~al.(2026)Yang, Wu, Wang, et~al.]{yang2026mineruchem}
H.~Yang, J.~Wu, J.~Wang, et~al.
\newblock {MinerU.Chem}: A high-precision system for optical chemical structure and reaction recognition, 2026.
\newblock URL \url{https://arxiv.org/abs/2608.03525}.

\bibitem[Yuan et~al.(2024)Yuan, Gong, and Xu]{yuan2024zenodo}
Shen Yuan, Shukai Gong, and Hongteng Xu.
\newblock {USPTO-LLM}: A large language model-assisted information-enriched chemical reaction dataset, 2024.
\newblock URL \url{https://zenodo.org/records/14396156}.

\bibitem[Yuan et~al.(2025)Yuan, Gong, and Xu]{yuan2025usptollm}
Shen Yuan, Shukai Gong, and Hongteng Xu.
\newblock {USPTO-LLM}: A large language model-assisted information-enriched chemical reaction dataset.
\newblock In \emph{Companion Proceedings of the ACM Web Conference 2025}, pages 817--820, 2025.
\newblock \doi{10.1145/3701716.3715295}.

\end{thebibliography}

\clearpage
\beginappendix
\section{Data snapshot, object definitions, and counting conventions}
\label{app:data-snapshot-definitions}

\setcounter{table}{0}
\renewcommand{\thetable}{A\arabic{table}}

This appendix records the data snapshot, core object definitions, and counting conventions used in this report.
Unless otherwise stated, the database-scale statistics reported in the main text and this appendix are based on the data snapshot dated \textbf{2026-06-25}, with statistics computed as of \textbf{2026-06-25}.

\subsection{Core data objects and terminology}
\label{app:core-data-objects}

The counts of \dsReferences{}, \dsSubstances{}, \dsReactionInstances{}, \dsReactionGroups{}, and \dsReactionTemplates{} refer to different database objects and cannot be added together as a single reaction count.

\begin{table*}[t]
\centering
\footnotesize
\setlength{\tabcolsep}{2.5pt}
\renewcommand{\arraystretch}{0.9}
\caption{Core DianShi-RxnDB objects and terminology.}
\label{tab:core-object-definitions}
\begin{tabular}{p{3.2cm}p{5.7cm}p{6.1cm}}
\toprule
Object or term &
Definition &
Primary use \\
\midrule
\dsReference{} &
A source patent document represented as a database object &
Connects patent metadata and source content with \dsReactionInstances{} \\
\dsSubstance{} &
A substance record organized by chemical identity in the database &
Supports name, identifier, and structure retrieval; links to related reactions \\
\dsReactionInstance{} &
Specific single-step reaction record extracted from a patent and retained in the database &
Stores participant roles, conditions, yield, procedure, and provenance \\
\dsReactionGroup{} &
Groups \dsReactionInstances{} with the same normalized reactant--product identity &
Organizes instances for one normalized reactant--product combination \\
\dsReactionTemplate{} &
Abstract reaction representation using SMARTS &
Provides an abstract representation beyond a specific reactant--product combination \\
Reaction Participant &
A participation relationship between a \dsSubstance{} and a \dsReactionInstance{} &
Records a \dsSubstance{} role as Reactant, Product, Reagent, Solvent, or Catalyst \\
\bottomrule
\end{tabular}
\end{table*}

\subsection{Patent-record processing stages and counting conventions}
\label{app:patent-processing-counts}

The patent-related scale statistics in this report use three principal counting conventions, while the processing-stage counts are reported in \Cref{tab:patent-processing-stages}.

\begin{enumerate}
\item \textbf{20,256,438 source patent records.}
This is the pre-deduplication source-record pool formed during data acquisition.

\item \textbf{1,580,939 patent documents entering the reaction-information extraction pipeline.}
This is the extraction corpus formed after target-domain filtering, record consolidation, and full-text availability checks.

\item \textbf{608,309 \dsReferences{}.}
In the database snapshot used by this report, these are source patent documents linked to at least one retained \dsReactionInstance{}.
\end{enumerate}

These values correspond to different processing stages and are not interchangeable.

\section{Core object fields and reaction-process operation definitions}
\label{app:core-object-fields-operations}

\setcounter{table}{0}
\renewcommand{\thetable}{B\arabic{table}}

This appendix supplements the principal information organized for \dsReference{}, \dsReactionInstance{}, and \dsSubstance{} objects and lists the operational definitions used for reaction-process details.
Field availability depends on the applicable data record and service version.

\subsection{\dsReference{} fields}
\label{app:reference-fields}

\begin{table*}[t]
\centering
\small
\caption{Principal \dsReference{} fields.}
\label{tab:reference-fields}
\begin{tabular}{p{2.8cm}p{4.4cm}p{7.0cm}}
\toprule
Field &
Description &
Example \\
\midrule
Document name &
Patent name or title &
Therapeutic agent for Parkinson disease \\
Document type &
Type of source document &
\texttt{PATENT} \\
Document number &
Patent identifier &
\texttt{US054849209-19960116} \\
Authors or organization &
Patent inventors or organization &
Suzuki; Fumio; Shimada; Junichi; Koike; Nobuaki et al. \\
\bottomrule
\end{tabular}
\end{table*}

\subsection{\dsReactionInstance{} fields}
\label{app:reaction-instance-fields}

\begingroup
\footnotesize
\begin{longtable}{p{2.8cm}p{3.8cm}p{7.7cm}}
\caption{Principal \dsReactionInstance{} fields.}
\label{tab:reaction-instance-fields}\\
\toprule
Field or group &
Description &
Example (\href{https://dianshi.opendatalab.org.cn/reaction/3b4d28ab429fb94b779f6724}{DianShi-RxnDB Web record}) \\
\midrule
\endfirsthead
\toprule
Field or group &
Description &
Example (\href{https://dianshi.opendatalab.org.cn/reaction/3b4d28ab429fb94b779f6724}{DianShi-RxnDB Web record}) \\
\midrule
\endhead
In-document number &
Subheading at the provenance location of the reaction in the patent &
Intermediate 68 \\
Yield &
Product yield &
75\% \\
SMILES &
Simplified SMILES &
\nolinkurl{[CH3:1][c:2]1[cH:3][c:4]2[c:5]([cH:6][n:7][n:8]2[CH:9]2[CH2:10][CH2:11][CH2:12][CH2:13][O:14]2)[cH:15][c:16]1[N+:17](=O)[O-]>>[CH3:1][c:2]1[cH:3][c:4]2[c:5]([cH:6][n:7][n:8]2[CH:9]2[CH2:10][CH2:11][CH2:12][CH2:13][O:14]2)[cH:15][c:16]1[NH2:17]} \\
Confidence &
Confidence for the SMILES &
86.80\% \\
Condition: Temp. &
Reaction temperature &
RT \\
Condition: Time &
Reaction time &
18h \\
Condition: Pressure &
Reaction pressure &
1 atm \\
Condition: Atmos. &
Reaction atmosphere &
H2 \\
Role: Reactant &
Reactant &
6-methyl-5-nitro-1-tetrahydropyran-2-yl-indazole \\
Role: Product &
Product &
6-methyl-1-(tetrahydro-2H-pyran-2-yl)-1H-indazol-5-amine \\
Role: Catalyst &
Catalyst &
palladium on carbon \\
Role: Solvent &
Solvent &
ethyl acetate \\
Process details &
Stepwise operations and corresponding action descriptions &
\parbox[t]{7.7cm}{\footnotesize\ttfamily 1. [MakeSolution] A suspension of 6-methyl-5-nitro-1-tetrahydropyran-2-yl-indazole (1250 mg, 4.78 mmol) in ethyl acetate (12 mL) and palladium on carbon (0.03 g, 0.3 mmol) was vigorously stirred\\ 2. [Stir] stirred for 18 h at RT under 1 atm of hydrogen\\ 3. [Filter] The mixture was filtered over Celite\textsuperscript{TM}\\ 4. [Concentrate] The filtrate was concentrated under reduced pressure\\ 5. [Purify] purified by silica column chromatography eluting with 10--70\% EtOAc in Pet. Ether\\ 6. [Triturate] which was triturated with diethyl ether to give $-1$ (833 mg, 3.6 mmol, 75\% yield) as a beige solid} \\
Workup details &
Workup descriptions such as concentration and purification &
The mixture was filtered over Celite\textsuperscript{TM}. The filtrate was concentrated under reduced pressure and purified by silica column chromatography eluting with 10--70\% EtOAc in Pet. Ether to give a brown oil which was triturated with diethyl ether to give 6-methyl-1-tetrahydropyran-2-yl-indazol-5-amine (833 mg, 3.6 mmol, 75\% yield) as a beige solid. \\
Provenance text &
Detailed description associated with the reaction in the source patent &
A suspension of 6-methyl-5-nitro-1-tetrahydropyran-2-yl-indazole (1250 mg, 4.78 mmol) in EtOAc (12 mL) and 10\% palladium on carbon (dry, 0.03 g, 0.3 mmol) was vigorously stirred for 18 h at RT under 1 atm of H2. The mixture was filtered over Celite\textsuperscript{TM}. The filtrate was concentrated under reduced pressure and purified by silica column chromatography eluting with 10--70\% EtOAc in Pet. Ether to give a brown oil which was triturated with diethyl ether to give 6-methyl-1-tetrahydropyran-2-yl-indazol-5-amine (833 mg, 3.6 mmol, 75\% yield) as a beige solid. \\
\bottomrule
\end{longtable}
\endgroup

\subsection{\dsSubstance{} fields}
\label{app:substance-fields}

\begingroup
\small
\begin{longtable}{p{3.3cm}p{4.1cm}p{6.9cm}}
\caption{Principal \dsSubstance{} fields and association statistics.}
\label{tab:substance-fields}\\
\toprule
Field or group &
Description &
Example \\
\midrule
\endfirsthead
\toprule
Field or group &
Description &
Example \\
\midrule
\endhead
\dsSubstance{} name &
IUPAC name of the substance &
8-(E)-3,4-Dimethoxystyryl-1,3-dipropyl-7-methylxanthine \\
Molecular formula &
Molecular formula of the substance &
\texttt{C22H28N4O4} \\
Molecular weight &
Molecular weight of the substance &
412.49 \\
Canonical SMILES &
Canonical SMILES of the substance &
\nolinkurl{CCCn1c(=O)c2c(nc(/C=C/c3ccc(OC)c(OC)c3)n2C)n(CCC)c1=O} \\
InChI &
International Chemical Identifier &
\nolinkurl{1S/C22H28N4O4/c1-6-12-25-20-19(21(27)26(13-7-2)22(25)28)24(3)18(23-20)11-9-15-8-10-16(29-4)17(14-15)30-5/h8-11,14H,6-7,12-13H2,1-5H3/b11-9+} \\
Association statistics for 1,4-butanediol &
Patent records containing the \dsSubstance{} &
2,302 \\
 &
\dsReactionInstances{} containing the \dsSubstance{} &
5,475 \\
 &
Appearances as Product &
236 \\
 &
Appearances as Reactant &
5,026 \\
 &
Appearances as Reagent &
269 \\
 &
Appearances as Solvent &
279 \\
 &
Appearances as Catalyst &
23 \\
\bottomrule
\end{longtable}
\endgroup

\subsection{Reaction-process operation definitions}
\label{app:reaction-operation-definitions}

Reaction-process details record operation types and the corresponding action descriptions step by step.
The current reference definition contains 38 operation types.

\begingroup
\small
\begin{longtable}{p{3.8cm}p{10.8cm}}
\caption{Operational definitions for reaction-process details.}
\label{tab:reaction-operation-definitions}\\
\toprule
Operation &
Definition \\
\midrule
\endfirsthead
\toprule
Operation &
Definition \\
\midrule
\endhead
Add & Add a substance to the reactor \\
Centrifuge & Centrifuge the mixture to separate phases or collect solids \\
CollectLayer & Select aqueous or organic fraction(s) \\
Column & Execute column chromatography using adsorbent and eluent \\
Combine & Combine multiple batches, fractions, or extracts \\
Concentrate & Evaporate the solvent (e.g., using a rotary evaporator) \\
Crush & Crush or grind solid material (e.g., with mortar and pestle) \\
Decant & Decant the supernatant liquid from a solid or separate layers \\
Degas & Purge the reaction mixture with a gas \\
Dilute & Dilute a solution by adding solvent \\
Distill & Separate or purify components by distillation \\
DrySolid & Dry a solid \\
DrySolution & Dry an organic solution with a desiccant \\
Evaporate & Remove solvent through evaporation, often under reduced pressure \\
Extract & Transfer compound into a different solvent \\
Filter & Separate solid and liquid phases \\
Freeze & Freeze the solution or mixture (e.g., before lyophilization) \\
Lyophilize & Freeze-dry or lyophilize the solution to remove solvent \\
MakeSolution & Mix several substances to generate a mixture or solution \\
Microwave & Heat the reaction mixture in a microwave apparatus \\
Partition & Add two immiscible solvents for subsequent phase separation \\
PH & Change the pH of the reaction mixture \\
PhaseSeparation & Separate the aqueous and organic phases \\
Precipitate & Precipitate a solid from solution \\
Purify & Perform purification (typically chromatography) \\
Quench & Stop reaction by adding a substance \\
Recrystallize & Recrystallize a solid from a solvent or mixture of solvents \\
Reflux & Reflux the reaction mixture \\
SetTemperature & Change the temperature of the reaction mixture \\
SetPressure & Adjust the pressure of the system (e.g., for distillation or evaporation) \\
Sonicate & Agitate the solution with sound waves \\
Stir & Stir the reaction mixture for a specified duration \\
Transfer & Transfer a reagent or mixture between vessels \\
Triturate & Triturate the residue \\
Wait & Leave the reaction mixture to stand for a specified duration \\
Wash & Wash after filtration or with an immiscible solvent \\
Yield & Record product information such as composition, appearance, mass, or concentration \\
OtherAction & Handle text that does not correspond to the defined actions above \\
\bottomrule
\end{longtable}
\endgroup

\section{Manual quality-evaluation protocol and statistical methods}
\label{app:quality-evaluation-protocol}

\setcounter{table}{0}
\renewcommand{\thetable}{C\arabic{table}}

\subsection{Evaluation population and scope}
\label{app:evaluation-population}

The evaluation used the 14,808,205 qualified \dsReactionInstances{} as its target population and randomly sampled 1,300 reaction records from that population.
Each sampled record was evaluated for five fields---Yield, Reactant, Reagent, Catalyst, and Solvent---yielding 6,500 field-level judgments.
The resulting estimates apply only to the qualified-instance population in the evaluated data batch and do not estimate the quality of non-qualified instances, unevaluated fields, database recall, or the correctness of an entire \dsReactionInstance{}.

\subsection{Annotation, relabeling, and review procedure}
\label{app:annotation-review}

The quality-evaluation process included independent annotation,
disagreement-triggered re-annotation, and consolidation of the final labels.
Each reaction record was independently evaluated by two annotators, both of
whom assessed all five fields using the structured record and the corresponding
source patent. If the two annotators disagreed on at least one of the five
fields, the entire record was independently re-evaluated by a third annotator.
For these records, the final label for each field was determined by majority
vote across the three annotations. The reported statistics were calculated
using the final labels obtained after this process.

\subsection{Statistical definitions}
\label{app:statistical-definitions}

For each evaluated field, field-level accuracy is the number of correct judgments divided by the number of evaluated records for that field:

\begin{equation}
\operatorname{Accuracy}_{f}
=
\frac{C_f}{N_f},
\qquad N_f = 1{,}300,
\label{eq:field-accuracy}
\end{equation}

where $C_f$ is the number of correct judgments for field $f$ and $N_f$ is that field's evaluated-record denominator.
The five-field micro-averaged accuracy pools all correct field-level judgments and all field-level denominators:

\begin{equation}
\operatorname{MicroAccuracy}
=
\frac{\sum_f C_f}{\sum_f N_f}
=
\frac{\sum_f C_f}{5 \times N_f}
=
\frac{6{,}042}{6{,}500}
=
92.95\%.
\label{eq:micro-accuracy}
\end{equation}

The term \emph{accuracy} in this report denotes the proportion of field-level binary judgments marked correct under the manual quality-evaluation procedure.

The reported micro-average is a pooled field-level accuracy, not a record-level accuracy or an entity-matching precision, recall, or F1 score.

\section{External comparison with the Pistachio Reaction Dataset}
\label{app:pistachio-comparison}

\setcounter{table}{0}
\renewcommand{\thetable}{D\arabic{table}}

This appendix documents the protocol and supporting evidence for the external matched comparison reported in \Cref{sec:pistachio-comparison}, covering reaction-record scale after deduplication, field richness and representation granularity, and field-level exact agreement against source-grounded references.
The comparison treated Pistachio as an external comparator rather than as ground truth and evaluated the two datasets within aligned samples without attempting to reconstruct either database's complete internal schema.

\subsection{Comparison scope and data release}
\label{app:pistachio-protocol}
\label{app:pistachio-objectives}

Record count, representational richness, and field-level agreement were evaluated separately because they describe different properties and should not be collapsed into a single score.
Although the vendor's website documented a 2026Q2 release at the time of writing, all record-level Pistachio analyses in this appendix used the 2025Q2 release available to us.
The evaluated Pistachio records span patent publication years from 1971 through 2025, whereas the DianShi-RxnDB corpus primarily covers patents published from 1976 through 2025; the comparison therefore does not assume identical literature coverage.

\subsection{Reaction-record scale after deduplication}
\label{app:pistachio-scale}

\subsubsection{Pistachio channels and the motivation for deduplication}
\label{app:pistachio-channels}

The sampling frame comprised US patents for which both datasets contained extraction results; 100 patents were randomly selected, and the scale analysis used all associated reaction records.

In the Pistachio records examined in this study, reaction records were supplied through two channels: a text channel extracted from textual reaction descriptions and an image channel extracted from reaction schemes or other graphical content.
The same patent example could therefore be represented by separate text-channel and image-channel records.
These channel-specific records could have different record identifiers, middle agent sections, role labels, names, or paragraph-text fields while still describing the same underlying reactant-to-product transformation.

If the two channels were counted independently, one underlying patent reaction could contribute more than one record to the Pistachio total.
Directly comparing such channel-combined counts with DianShi-RxnDB would therefore conflate reaction-record scale with overlapping channel representations.

\subsubsection{Illustrative text--image overlap in a patent}

A representative example is provided by patent US07795244.
The record identifiers, source-text availability, and deduplication result are reported in \Cref{tab:app-pistachio-us07795244-records}.
The values in the table are reproduced from the two compared JSON records; the long paragraph-text value is shown from its beginning.

\begin{table*}[t]
\centering
\scriptsize
\setlength{\tabcolsep}{3pt}
\renewcommand{\arraystretch}{1.08}
\caption{Raw record identifiers, source text, and deduplication result for the text- and image-channel records of the same patent example in US07795244.}
\label{tab:app-pistachio-us07795244-records}
\begin{tabular}{p{2.6cm}p{5.6cm}p{5.6cm}}
\toprule
Raw JSON field & Text channel & Image channel \\
\midrule
\texttt{title} & \texttt{US07795244B2\_0103} & \texttt{US07795244\_C00043} \\
\texttt{data.paragraphText} & \texttt{In step 5, in a suitable reaction flask, sodium iodide (23.9 g, 159.45 mmol, 1.6 eq.) was dissolved in acetone (96 mL). 2-Ethoxy-isobutyric acid chloromethyl ester (18 g, 99.65 mmol, 1 eq.) was then added ...} & Field absent \\
Deduplication result & Same reactant and product sets after removing atom-map numbers and sorting components & Same reactant and product sets after removing atom-map numbers and sorting components \\
\bottomrule
\end{tabular}
\end{table*}

The complete raw reaction-SMILES values were:

\noindent\textbf{Text channel, \texttt{data.smiles}}
\begin{lstlisting}[basicstyle=\ttfamily\scriptsize,breaklines=true,breakatwhitespace=false,columns=fullflexible]
[CH3:1][CH2:2][O:3][C:4]([CH3:5])([CH3:6])[C:7](=[O:8])[O:9][CH2:10]Cl.[Na][I:11]>CC(=O)C.N#N>[CH3:1][CH2:2][O:3][C:4]([CH3:5])([CH3:6])[C:7](=[O:8])[O:9][CH2:10][I:11]
\end{lstlisting}

\noindent\textbf{Image channel, \texttt{data.smiles}}
\begin{lstlisting}[basicstyle=\ttfamily\scriptsize,breaklines=true,breakatwhitespace=false,columns=fullflexible]
[CH3:1][CH2:2][O:3][C:4]([CH3:5])([CH3:6])[C:7](=[O:8])[O:9][CH2:10]Cl.[Na][I:11]>CC(=O)C>[CH3:1][CH2:2][O:3][C:4]([CH3:5])([CH3:6])[C:7](=[O:8])[O:9][CH2:10][I:11]
\end{lstlisting}

The selected raw \texttt{components} values were:

\noindent\textbf{Text channel}
\begin{lstlisting}[basicstyle=\ttfamily\scriptsize,breaklines=true,breakatwhitespace=false,columns=fullflexible]
[{'role':'Product','name':'2-ethoxy-isobutyric acid iodomethyl ester','smiles':'ICOC(C(C)(C)OCC)=O'}, {'role':'Reactant','name':'sodium iodide','smiles':'[Na]I'}, {'role':'Reactant','name':'2-Ethoxy-isobutyric acid chloromethyl ester','smiles':'ClCOC(C(C)(C)OCC)=O'}, {'role':'Solvent','name':'acetone','smiles':'CC(=O)C'}, {'role':'Solvent','name':'acetone','smiles':'CC(=O)C'}, {'role':'Agent','name':'nitrogen atmosphere','smiles':'N#N'}]
\end{lstlisting}

\noindent\textbf{Image channel}
\begin{lstlisting}[basicstyle=\ttfamily\scriptsize,breaklines=true,breakatwhitespace=false,columns=fullflexible]
[{'role':'Reactant','smiles':'O(C(C(OCC)(C)C)=O)CCl'}, {'role':'Agent','smiles':'[Na]I'}, {'role':'Agent','smiles':'CC(=O)C'}, {'role':'Product','smiles':'O(C(C(OCC)(C)C)=O)CI'}]
\end{lstlisting}

The reactant and product sections are identical after atom-map removal and component sorting; only the middle agent section differs, with \texttt{N\#N} appearing only in the text-channel reaction SMILES.
The text-channel record contains paragraph text and more detailed role and name information, while the image-channel record lacks the paragraph-text field and assigns sodium iodide and acetone to the broad \texttt{Agent} role.
The two channel-specific records therefore represent one underlying reaction under the within-patent deduplication rule described below.

\subsubsection{Within-patent deduplication rule}

To make the reaction-record comparison more comparable across the two resources, deduplication was performed within each patent using a common reactant--product-based rule.
For both datasets, two records within the same patent were treated as duplicates only when their reactant and product sets were identical after atom-map removal, component normalization, and sorting.
Component order and the middle agent section were ignored, while differences in agents, source text, identifiers, names, or participant roles alone did not prevent duplicate classification.

\subsubsection{Matched-sample results}
\label{app:pistachio-deduplication}

The 638 removed Pistachio duplicates comprised 115 text-internal duplicates, 89 image-internal duplicates, and 434 text--image overlaps.
After deduplication, DianShi-RxnDB retained 4,093 of 4,222 records and Pistachio retained 2,992 of 3,630 records, giving a DianShi-RxnDB/Pistachio ratio of 1.368 in the matched sample (\Cref{tab:pistachio-deduplication}).
DianShi-RxnDB contained more retained records in 51 patents, Pistachio in 42, and the remaining 7 patents were tied.

\subsection{Field richness and representation granularity}
\label{app:pistachio-representation}

This comparison inspected the fields and objects exposed for the shared reaction \texttt{Intermediate 4} in patent US10975080; both records describe oxidation of the same alcohol to the same ketone.
\Cref{tab:app-pistachio-representation} reproduces the observed values, with a dash indicating that the inspected Pistachio record did not expose the item as a separate field; this record-level evidence is not a claim about all records or other parts of the Pistachio product.

\begin{table*}[t]
\centering
\scriptsize
\setlength{\tabcolsep}{3pt}
\renewcommand{\arraystretch}{1.08}
\caption{Concrete field-level evidence for the shared reaction \texttt{Intermediate 4} in patent US10975080.}
\label{tab:app-pistachio-representation}
\begin{tabular}{p{3.0cm}p{5.7cm}p{5.7cm}}
\toprule
Field in this reaction & DianShi-RxnDB & Pistachio \\
\midrule
Source pointer & \texttt{Intermediate 4}; source line 1306; full paragraph & \texttt{US10975080B2\_0290}; \texttt{paragraphText} \\
Role labels & Reactant; Reagent; Solvent; Product & Reactant; broad \texttt{Agent}; Solvent; Product \\
Catalyst field & Explicitly recorded and empty for this reaction & --- No separate Catalyst field \\
Reaction stage & \texttt{stage\_1} groups substances, conditions, actions, and workup & --- No stage-level object \\
Temperature & Stage field: \texttt{0~\textdegree C} & Action parameter: \texttt{Temperature = 0~\textdegree C} \\
Time & Stage field: \texttt{2 h} & Action parameter: \texttt{Time = 2 h} \\
Ordered procedure & 10 typed actions with IDs & 10 action records with text \\
Workup & Dedicated workup field & --- No dedicated workup field \\
Product and amount & Product name and SMILES; \texttt{470 mg}; \texttt{1.84 mmol}; derived \texttt{87.2\%} yield & Product and SMILES; typed quantities, including derived yield \\
Structure and name normalization & Source wording, normalized name, standard SMILES, and structure source & Component name, SMILES, and InChI \\
Reaction validation & SMILES confidence, missing-atom check, ratios, and reaction scale & --- No equivalent diagnostics in this record \\
\bottomrule
\end{tabular}
\end{table*}

DianShi-RxnDB exposes finer-grained participant roles and more independently retrievable process, workup, provenance, and validation dimensions in this inspected shared-reaction example.

\subsection{Field-level exact agreement against source-grounded references}
\label{app:pistachio-quality}

Starting from candidate records in the 100 sampled patents, we grouped records by patent number and retained a pair only when its source paragraphs from the patent XML/SGM were identical after lowercasing and removal of markup, punctuation, whitespace, and line-break differences.
This procedure yielded 660 source-paragraph-matched reaction pairs from 58 patents.
The matched paragraphs established a shared source passage but did not establish extraction correctness or completeness, or recover entities omitted by both datasets.
They provided the evidence for constructing source-grounded reference sets: candidate entities identified by either dataset could be considered, workup-only entities were excluded from reaction-participant references, and Pistachio was not treated as ground truth.

The comparison covered Reactant, Reagent, Catalyst, Solvent, Product, and Yield.
Because Pistachio uses a broad \texttt{Agent} category, both datasets were mapped to a shared role vocabulary: source-supported catalysts and reaction media were assigned to Catalyst and Solvent, respectively; hydrogen used as a reaction input or reducing agent was assigned to Reagent; and workup-only entities, filtration aids, and atmospheric gases were excluded.
Names and common aliases were normalized before comparison.
Genuine role ambiguities could be treated as optional, while Reactant used a stricter direct-role audit so that explicit omissions, additions, and role errors remained errors.

GPT-5.6-sol performed LLM-assisted semantic adjudication under a fixed extraction-field prompt based on the source patent evidence and field definitions; deterministic code performed record alignment, normalization, set construction, field-level agreement assessment, counting, and percentage calculation.
Each matched reaction contributed one binary observation per field.
A field was counted as correct only when its complete normalized extracted field set exactly matched the source-grounded reference set; missing or extra items were counted as errors, and legitimately empty fields remained in the denominator.
This metric is field-level exact agreement rather than substance-level precision, recall, or F1, or whole-record accuracy; structure-level correctness and atom mapping were outside its scope.

The resulting values are reported in \Cref{tab:pistachio-field-comparison}.
For Yield, DianShi-RxnDB generally records an explicit percentage when present in the source text, whereas Pistachio may provide a value inferred from product mass and stoichiometry when no percentage is stated.
Under the source-text exactness criterion used here, an inferred value was not counted as an explicit percentage, so the Yield difference reflects both extraction behavior and representation policy.

\subsection{Limitations}
\label{app:pistachio-limitations}

The deduplication results describe the matched sample of 100 US patents, while the field-level results are based on 660 source-paragraph-matched reaction pairs from 58 patents; neither was extrapolated to the complete corpora.
The field-level comparison is affected by source-grounded reference construction, role harmonization, ambiguity treatment, and Yield representation policy, and should not be pooled with the internal evaluation of 1,300 qualified DianShi-RxnDB records; the representation comparison is limited to the inspected shared-reaction example.

\end{document}